\documentclass[Afour,sageh,times]{sagej}

\usepackage{moreverb,url}

\usepackage[colorlinks,bookmarksopen,bookmarksnumbered,citecolor=red,urlcolor=red]{hyperref}

\usepackage{graphicx} 
\usepackage{amsmath} 
\usepackage{amssymb}  
\usepackage{bm}
\usepackage{multirow}
\usepackage{xcolor}
\usepackage{pifont}
\usepackage{caption}
\usepackage{subcaption}
\usepackage{hyperref}
\newtheorem{definition}{Definition}
\usepackage{esint}
\usepackage{tikz}
\usepackage{wrapfig}
\usepackage[linesnumbered,ruled,norelsize]{algorithm2e}

\SetCommentSty{mycommfont}
\renewcommand{\b}[1]{\boldsymbol{#1}}

\newcommand{\hh}[1]{\textcolor{black}{#1}} 
\newcommand\bluesout{\bgroup\markoverwith{\textcolor{blue}{\rule[0.5ex]{2pt}{0.4pt}}}\ULon}

\usepackage[normalem]{ulem}

\newcommand\BibTeX{{\rmfamily B\kern-.05em \textsc{i\kern-.025em b}\kern-.08em
T\kern-.1667em\lower.7ex\hbox{E}\kern-.125emX}}

\def\volumeyear{2024}
\begin{document}

\runninghead{Martinez-Baselga et al.}

\title{SHINE: Social Homology Identification for Navigation in Crowded Environments}

\author{Diego Martinez-Baselga\affilnum{1}, Oscar de Groot\affilnum{2}, Luzia Knoedler\affilnum{2}, Luis Riazuelo\affilnum{1}, Javier Alonso-Mora\affilnum{2}, Luis Montano\affilnum{1}}

\affiliation{\affilnum{1}University of Zaragoza, Spain\\
\affilnum{2}Delft University of Technology, Netherlands}

\corrauth{Diego Martinez-Baselga, Robotics, Computer Vision and Artificial Intelligence Group, Institute of Engineering Research (I3A), University of Zaragoza, 50018 Zaragoza, Spain.}

\email{diegomartinez@unizar.es}

\begin{abstract}
Navigating mobile robots in social environments remains a challenging task due to the intricacies of human-robot interactions. Most of the motion planners designed for crowded and dynamic environments focus on choosing the best velocity to reach the goal while avoiding collisions, but do not explicitly consider the high-level navigation behavior (avoiding through the left or right side, letting others pass or passing before others, etc.). In this work, we present a novel motion planner that incorporates topology distinct paths representing diverse navigation strategies around humans. The planner selects the topology class that imitates human behavior the best using a deep neural network model trained on real-world human motion data, ensuring socially intelligent and contextually aware navigation. Our system refines the chosen path through an optimization-based local planner in real time, ensuring seamless adherence to desired social behaviors. In this way, we decouple perception and local planning from the decision-making process. We evaluate the prediction accuracy of the network with real-world data. In addition, we assess the navigation capabilities in both simulation and a real-world platform, comparing it with other state-of-the-art planners. We demonstrate that our planner exhibits socially desirable behaviors and shows a smooth and remarkable performance.
\end{abstract}


\newcommand\copyrighttext{%
  \footnotesize \textcopyright This paper has been accepted for publication at The International Journal of Robotics Research. Please, when citing the paper, refer to the official manuscript with the following DOI: 10.1177/02783649251344639.}
\newcommand\copyrightnotice{%
\begin{tikzpicture}[remember picture,overlay]
\node[anchor=south,yshift=10pt] at (current page.south) {\fbox{\parbox{\dimexpr\textwidth-\fboxsep-\fboxrule\relax}{\copyrighttext}}};
\end{tikzpicture}%
}

\maketitle


\section{Introduction} \label{sec:intro}

Some of the most important applications of mobile robots involve moving alongside humans. However, their seamless integration into social settings presents notable challenges. These challenges include aspects such as safety, efficiency, smooth interactions, adherence to social conventions, and gaining human acceptance. Addressing them is essential for the successful integration of robots into social spaces.

\begin{figure}[ht]
    \centering
    \includegraphics[width=0.55\linewidth]{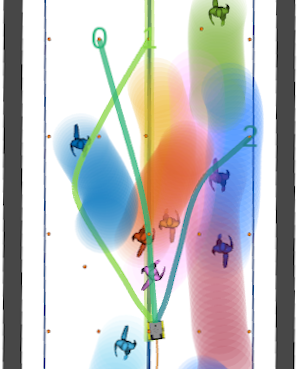}
    \caption{A dense corridor scenario, where the robot may follow one of three high level behaviors to progress. The behaviors are represented with three possible paths, and the future trajectories of humans with semi transparent shadows.}
    \label{fig:motivation}
\end{figure}


Motion planning in dynamic environments involves global and local planning. Traditional global planners compute a global plan that only considers static obstacles and a local planner tries to follow it. Thus, the local planner may find itself stuck in locally optimal plans that overlook the presence of dynamic obstacles beyond its planning horizon. This can lead to different dynamic behaviors that were not considered by the initial plan, such as passing before obstacles or letting them pass.

In addition, social navigation poses the challenge of selecting the desired robot dynamic behavior. While manually defining a cost based on heuristics is an option, evaluating social norms and adapting to every possible situation is a very complex problem. For example, in Figure~\ref{fig:motivation}, the robot faces a very dense scenario where it has to make a high-level decision, avoid and cooperate with humans to advance in a corridor. It may wait for the humans that are in front of it to move away or navigate through the left or right sides, dealing with the humans that it encounters there. In other scenarios, those kind of decisions could include aspects such as avoiding uncomfortable situations, not blocking other's way or understanding preferences.

This work aims to develop a motion planner, which we name \textbf{SHINE} (\textbf{S}ocial \textbf{H}omology \textbf{I}dentification for \textbf{N}avigation in crowded \textbf{E}nvironments), that uses high-level navigation decisions learned from humans for navigating in crowded environments. Out of the different trajectories that a robot can follow to reach its goal, some exhibit similar topological characteristics. High-level decisions are understood as the possible navigation topology classes of the trajectories with respect to the surrounding humans of the environment. They also represent different feasible local optimal routes to a goal when considering robotic navigation as an optimization problem. The proposed planner first computes topology distinct guidance trajectories using a Visibility-Probabilistic Road Maps (PRM) method \citep{de2023globally}, and 
chooses one 
based on a model that replicates human selection. We use a deep neural network model trained in a real-world dataset of human motion to estimate which is the most probable behavior to be chosen by a human. Finally, the guidance trajectory is locally optimized with a Local Model Predictive Contouring Controller (LMPCC) \citep{brito2019model} that ensures that the robot follows the desired social high-level behavior. The high-level behavior is dynamically updated to accommodate for unexpected changes in the predicted future trajectories of the surrounding pedestrians.

We extensively evaluate and analyze our proposed method in three different ways. First, we measure the accuracy of the supervised learning approach to predict the human behavior, and qualitatively examine some of the predictions. Second, we assess the behavior of our method with quantitative experiments in simulation and compare it with other state-of-the-art social motion planners. Finally, we conduct qualitative experiments on a real ground platform to evaluate and compare SHINE against state-of-the-art social motion planners in crowded and challenging real-world scenarios.

The article is organized as follows. The context and contribution of the work is studied in Section~\ref{sec:related}. Section~\ref{sec:problem} analyzes the problem to be solved, Section~\ref{sec:pred} presents and studies the prediction framework and Section~\ref{sec:system} the navigation system. Section~\ref{sec:exp} and Section~\ref{sec:discussion} examine the evaluation performed and, lastly, Section~\ref{sec:conclusion} concludes the work.
\section{Related work and contribution}\label{sec:related}

\subsection{Social navigation}

\hh{Navigating in environments populated by humans poses a challenge that has received significant attention from the robotics community, analyzed in recent survey papers as \citet{mavrogiannis2023core} or \citet{singamaneni2024survey}. Early obstacle avoidance} approaches primarily focused on static obstacles, such as the dynamic window approach proposed in \citet{fox1997dynamic}, later improved by including moving obstacles in \citet{missura2019predictive}. Velocity obstacles, introduced in \citet{fiorini1998motion}, laid the foundation for numerous studies, including \citet{wilkie2009generalized} or \citet{lorente2018model}, which used the concept for modeling obstacle motion. Additionally, strategies like the Optimal Reciprocal Collision Avoidance (ORCA) \citep{van2011reciprocal} and other methods based on it like NH-ORCA \citep{alonso2012reciprocal} or VR-ORCA \citep{guo2021vr} have explored the principle of reciprocity in collision avoidance. They consider a shared responsibility between the robot and dynamic obstacles, facilitating collaborative maneuvers. In \citet{lorente2018model}, an evolution of the velocity obstacles is presented, where strategies of passing before and after obstacles are applied to reach maximum speed and minimum time to reach the goal.

The motion planners \hh{in} previously mentioned work \hh{use} a local view of the environment in a short spatial horizon, which may lead to local traps and non-smooth trajectories. In \citet{martinez2023SDOVS}, the planner developed in \citet{lorente2018model} for dynamic environments is integrated with a global planner, avoiding trap situations. Some approaches use a Model Predictive Controller (MPC) to compute optimal control commands for the robot in advance. For example, LMPCC \citep{brito2019model} is a method for real-time navigation among dynamic obstacles. Crowd motion estimation may be included in the optimizer to make it more reliable \citep{poddar2023crowd}.

The above approaches do not account for predictive capabilities of pedestrians, which can result in unsafe or unnatural behaviors \citep{kretzschmar2016socially}. Some methods include social constraints or try to imitate human motion. One example is the early method Social Forces \citep{helbing1995social} and its extended versions \citep{jiang2017extended}, which intend to model social interactions as a sum of attractive and repulsive forces. 

\hh{Previous methods typically treat other agents as dynamic obstacles whose trajectories are predicted, without fully addressing the complexities of interactions between decision-making agents. This may lead to defensive and opaque behaviors~\citep{trautman2015robot}. Thus, another important body of work addresses multi-agent interactions by coupling prediction and planning. Some methods do it explicitly using, for example, intent inference~\citep{bandyopadhyay2013intention,bai2015intention}, a game-theoretic perspective~\citep{fridovich2020efficient,sun2021move,peters2024contingency,mehr2023maximum} or topological representations~\citep{mavrogiannis2017socially, mavrogiannis2019multi}. Since each agent’s action depends on the decisions of the others, explicitly solving the coupled prediction and planning problem presents a considerable computational challenge.}

Other approaches \hh{implicitly accounts for interactions with crowds in their design}. Particularly, deep reinforcement learning (DRL) planners show promising results \hh{by learning from interactions with the environment}. A common approach is using decentralized end-to-end algorithms whose output is the velocity of the robot \citep{chen2017decentralized}. Some of the networks use different layers to encode the environment and extract features to navigate, such as LSTM layers \citep{everett2018motion}, \citep{everett2021collision}, attention \citep{chen2019crowd} or graph neural networks \citep{chen2020relational},  \citep{chen2020robot}. However, navigating with humans require special considerations, due to social norms and special needs in collaboration and safety. In \citet{hu2022crowd}, social stress is introduced to the state space of the DRL framework, while in \citet{zhou2022robot} and \citet{martinez2023improving}, social attention is used to model the interactions between the humans and the robot. Nevertheless, DRL approaches have the limitations of being very dependent on the simulator and the modeling of people and robot interactions, as well as a lack of explicit collision constraints.

\subsection{Topology-based navigation}

Reactive planners focus on the immediate surroundings of the robot, while MPC-based planners optimize their current trajectory, leading to suboptimal paths. 
\hh{Two trajectories that start and end at the same points but cannot be transformed into one another without crossing an obstacle adopt different spatial configurations to avoid the obstacle.}We say that they are topological distinct trajectories. Homotopy classes refer to different topological paths or trajectories that are considered equivalent in terms of their spatial relationship (further explained in Section~\ref{sec:pred}). By computing trajectories in distinct homotopy classes, robots can choose trajectories that avoid obstacles and optimize their movements when unexpected obstacles or changes in the environment occur \citep{kolur2019online}. Topological information can be useful not only for navigation, but also for other tasks as information gathering \citep{mccammon2021topological}. Discovering topologically distinct trajectories can be done with Voronoi diagrams \citep{rosmann2017integrated}, but a more common approach in the literature are Probabilistic Roadmaps (PRM) \citep{simeon2000visibility} or derivatives \citep{novosad2023ctoprm}, which is a sampling-based method to construct a sparse roadmap.

Some works use the idea of choosing one of the topologically distinct trajectories as a guidance of an MPC framework, producing a locally optimal trajectory that is topologically equivalent to the guidance. This has enabled fast and safe flights to UAVs with perception-aware trajectory replanning \citep{zhou2021raptor}, and demonstrated state-of-the-art performance for flights in cluttered environments \citep{penicka2022learning}, \citep{penicka2022minimum}. 

More related to our problem, \citet{de2023globally} presents an approach to navigate in dynamic environments by finding topologically distinct guidance trajectories in the dynamic space and use LMPCC to find a global optimal plan. Another interesting approach \citep{mavrogiannis2022winding} adds a cost in the MPC to encourage the robot to be topologically invariant in time. A recent approach \citep{ding2023prtirl} includes a homotopic cost in the generation of nodes of an Optimal Rapidly-exploring Random Trees (RRT$^\ast$) algorithm. Nevertheless, RRT$^\ast$ is a planner for static environments and computes a global path regardless the future of the environment. Similar to our idea, H-TEB \citep{rosmann2017online} selects a topology class based on a cost learned by human parameters and plans a trajectory in that class. However, they only consider a static environment, use a cost function with only three parameters, need to predict the classes for the full scenario and assume the other agents have a collaborative behavior. Therefore, their navigation performance is strongly related to the collaboration of other agents and their predictions, which could be inaccurate due to the static selection of a topology in a dynamic environment.

A comparison of the main features of our proposed method with the discussed motion planners in dynamic environments is shown in Table~\ref{tab:navigation}.

\begin{table*}[ht]
    \caption{A comparison of our navigation planner with other planners for dynamic environments. The social behavior column refers to social features included in the planner, social learning whether those features are learned, C.D.P. scenarios if the planner is tested in Crowded, Dynamic and Possibly non-cooperative scenarios, and global planning checks the ability of the planner to avoid local traps or replan a different path from the one previously followed. }
    \centering
    \begin{tabular}{|c|c|c|c|c|c|}
        \hline
        \multirow{2}{*}{\textbf{Planner}} & \textbf{Social} & \textbf{Social} & \textbf{C.D.P 
        } & \textbf{Global}   & \textbf{Open}\\
         & \textbf{behavior} & \textbf{learning} & \textbf{scenarios}  & \textbf{planning}   & \textbf{source}\\
        \hline
         LMPCC \citep{brito2019model}                &   &   & \ding{51} &   & \ding{51} \\
         Guidance-MPCC \citep{de2023globally}      &   &   & \ding{51} & \ding{51} & \ding{51} \\
         Social Force \citep{helbing1995social}    & \ding{51} &   & \ding{51} &   & \ding{51} \\
         Social DRL \citep{martinez2023improving}   & \ding{51} & \ding{51} & \ding{51} &   & \ding{51} \\
         T-MPC \citep{mavrogiannis2022winding}      & \ding{51} &   & \ding{51} &   & \ding{51} \\
         SCN \citep{mavrogiannis2017socially}       & \ding{51} & \ding{51} &   &   &   \\
         PRTIRL \citep{ding2023prtirl}              & \ding{51} & \ding{51} &   & \ding{51} &  \\
         H-TEB  \citep{rosmann2017online}           & \ding{51} & \ding{51} &   & \ding{51} &  \\
         SHINE 
         & \ding{51} & \ding{51} & \ding{51} & \ding{51} & \ding{51} \\
        \hline
    \end{tabular}

    \label{tab:navigation}
\end{table*}

\subsection{Human navigation prediction}

Understanding and predicting humans' motion is crucial for intelligent robots to navigate safely and interact in crowds, and there is a growing trend in research focused on predicting human trajectories \citep{rudenko2020human}. That knowledge could be used by the robot to use the same social navigation rules, ensuring safety, facilitating collaboration, adapting to social context and exhibiting natural human-robot interaction. Strategies include using social attention \citep{vemula2018social}, human-oriented transformers \citep{yuan2021agentformer} or solutions based on retrospective memory \citep{xu2022remember}. In \citet{SalzmannIvanovicEtAl2020}, the authors present a modular model that incorporates agents dynamics and heterogeneous data, which they extend in \citet{IvanovicHarrisonEtAl2023} with adaptive meta-learning, showing that the method is easily extensible. These methods prove their performance in public real-world datasets, such as the ETH/UCY dataset \citep{lerner2007crowds} \citep{pellegrini2009you}, which have data of pedestrians positions in 5 different scenarios.

A recent work, VOMP-h \citep{wakulicz2023topological}, predicts homotopy classes. It discusses the problems of trajectory prediction based on geometric representations, explaining the benefits of predicting homotopy classes instead of the trajectories. However, this approach is not directly applicable to navigation and assumes a static environment. Furthermore, it estimates the partial homotopy class of the next obstacle faced by the robot, so it is hard to assess the scalability in dynamic and cluttered scenarios. Previously cited H-TEB \citep{rosmann2017online} also predicts the homotopy class followed by the humans. However, it does it in the scenarios as if they were static, not taking into account the past or the future of the environment, and uses a sum of three hand-crafted terms to predict the cost. The weight of the terms is learned from demonstrations.

Other work that predicts the topology of the environment is SCN \citep{mavrogiannis2017socially}, where a network is trained to predict the permutations in the positions of people moving in a Social Force simulator, and includes a planner based on braids to select a braid that meets the predicted permutation. In Table~\ref{tab:learning-homologies}, we summarize the main differences between our approach and these two methods. While SCN only learns permutations, VOMP-h only predicts homotopy classes in static environments and is not translated to navigation commands.

\begin{table*}[ht]
    \caption{A comparison of our learning path topologies approach with others. The dynamic environments' column refers to including dynamic obstacles in the predictions, navigation if the predictions may be applied in navigation, real-world data if it was used for training, homotopy class if they predict homotopy classes, and sophisticated model if the prediction model to capture humans' preferences is more complex than a simple addition of a few terms.}
    \centering
    \begin{tabular}{|c|c|c|c|c|c|c|}
        \hline
        \multirow{2}{*}{\textbf{Algorithm}} & \textbf{Dynamic}  & \multirow{2}{*}{\textbf{Navigation}} & \textbf{Real-world} & \textbf{Homotopy} &  \textbf{Sophisticated} & \textbf{Open}\\
        & \textbf{environments} &  & \textbf{data} & \textbf{class} & \textbf{model} & \textbf{source}\\
        \hline
         SCN \citep{mavrogiannis2017socially}        & \ding{51} & \ding{51} &   &  & \ding{51} &  \\
         H-TEB \citep{rosmann2017online}             &  & \ding{51} & \ding{51} & \ding{51} & & \\
         VOMP-h \citep{wakulicz2023topological}      &  &   & \ding{51} & \ding{51} & \ding{51}&  \\
         SHINE 
         & \ding{51} & \ding{51} & \ding{51} & \ding{51} & \ding{51} & \ding{51} \\
        \hline
    \end{tabular}

    \label{tab:learning-homologies}
\end{table*}

\subsection{Contribution}

The two-key contributions of this work are:

\begin{enumerate}
    \item A new supervised learning approach, able to learn human navigation preferences in crowded environments. Our approach learns how different the topology class of a queried trajectory is from that which a human would choose. This allows us to distinguish between social and non-social high-level navigation decisions.
    \item A new planner that infers a social high-level navigation decision to initialize an optimization-based planner (LMPCC \citep{brito2019model}), and may reactively replan with new social decisions if the scenario behaves unexpectedly.
\end{enumerate}

We evaluate how well our model predicts the human high-level navigation decisions in real-world scenarios and compare it with other variations and hand-crafted cost selections, and we show that our approach scales well in different scenarios and environments. Then, we extensively compare our approach with other motion planners in dynamic environments in both simulation and the real world, showing their differences and limitations. Additionally, we validate the behavior of our proposed planner in specific challenging scenarios and among five pedestrians in a tight space in a laboratory experiment, showing learned socially acceptable behaviors.

The system will be released as open source upon acceptance.
\section{Problem formulation}\label{sec:problem} 
We consider scenarios in which a robot must navigate to a goal position $\boldsymbol{p}_{goal}=(x_{goal},y_{goal})$ while ensuring collision avoidance with dynamic obstacles, i.e. humans.
We model the robot's motion by the deterministic discrete-time non-linear dynamics:
\begin{equation}
    \boldsymbol{x}_{k+1} = f(\boldsymbol{x}_k, \boldsymbol{u}_k),
\end{equation}
where $\boldsymbol{x}_k \in \mathbb{R}^{n_x}$ is the state and $\boldsymbol{u}_k \in \mathbb{R}^{n_u}$ the control input of the system at time step $k$, and $n_x$ and $n_u$ are the state and input dimensions, respectively. The state of the robot at time $k$ contains its 2-D position $\boldsymbol{p}_k=(x_k, y_k) \in \mathbb{R}^2 $. 

We will use the superscript $T$ when referring to the collection of variables over $T$ discrete time steps. For example, $\boldsymbol{p}^{T}_k =[\boldsymbol{p}_{k-T}, \dots, \boldsymbol{p}_{k}]$ denotes the robot's positions over the past $T$ discrete time steps.

The robot and the dynamic obstacles are represented as disks with radius $r_{robot}$ and $r_{obs}$, respectively. The position of the obstacle $j$ at time $k$ is denoted as $\boldsymbol{o}_{j,k}=(x_{j,k},y_{j,k}) \in \mathbb{R}^2$ and the position of the set of $M$ obstacles at time $k$ is denoted as $\mathcal{O}_k=[\boldsymbol{o}_{0,k}, \dots,\boldsymbol{o}_{M,k}]$. We assume that the previous positions of the obstacles are known for a time horizon $T$, denoted as $\mathcal{O}^T_k=[\mathcal{O}_{k-T}, \dots, \mathcal{O}_{k}]$. The values of $M$ and $T$ are not fixed.

The state space that we consider for planning is composed by the workspace and time: $\mathcal{X}=\mathbb{R}^2\times[0,T]$. A trajectory is a continuous path through the state space: $\boldsymbol{\tau}: [0,1]\rightarrow\mathcal{X}$\hh{, and $\mathcal{X}_{traj}$ the set of all possible trajectories in $\mathcal{X}$. We denote $\boldsymbol{\tau}_1 \sqcup \boldsymbol{\tau}_2$ the the curve formed by $\boldsymbol{\tau}_1$ and $\boldsymbol{\tau}_2$ together, and $-\boldsymbol{\tau}$ the same trajectory $\boldsymbol{\tau}$ but in opposite orientation.} The collision-free space describes the state space without the space occupied by the obstacles.

\subsection{Homotopy and Homology Classes}
We are interested in the high-level behavior of trajectories in the collision-free space (i.e., how they pass the obstacles) so that we can learn this behavior from humans. The direction in which a trajectory passes obstacles is captured by \textit{homotopy} classes, a concept from topology.

\hh{\begin{definition}
\citep{bhattacharya2012topological}~(Homotopic Trajectories) Two trajectories, $\boldsymbol{\tau}_1:[0,1]\rightarrow\mathcal{X}$ and $\boldsymbol{\tau}_2:[0,1]\rightarrow\mathcal{X}$, connected by the same starting and ending point, are homotopic if they can be continuously deformed into each other without intersecting with any obstacle, while keeping the endpoints. Homotopic trajectories belong to the same homotopy class.
\end{definition}}

\begin{figure}
    \centering
    \begin{tabular}{c}
      \includegraphics[width=0.9\linewidth]{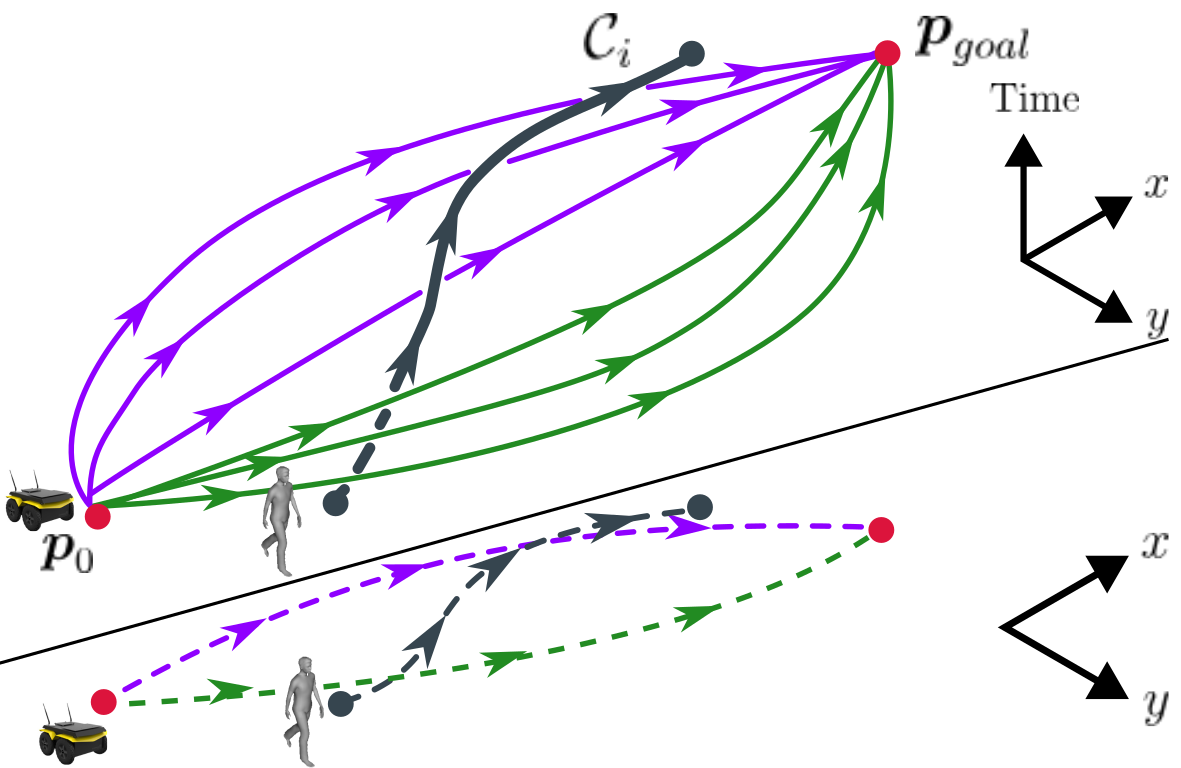}    \\
       \hh{\footnotesize{(a) Trajectories in different homotopy classes}}  \\
        \includegraphics[width=0.9\linewidth]{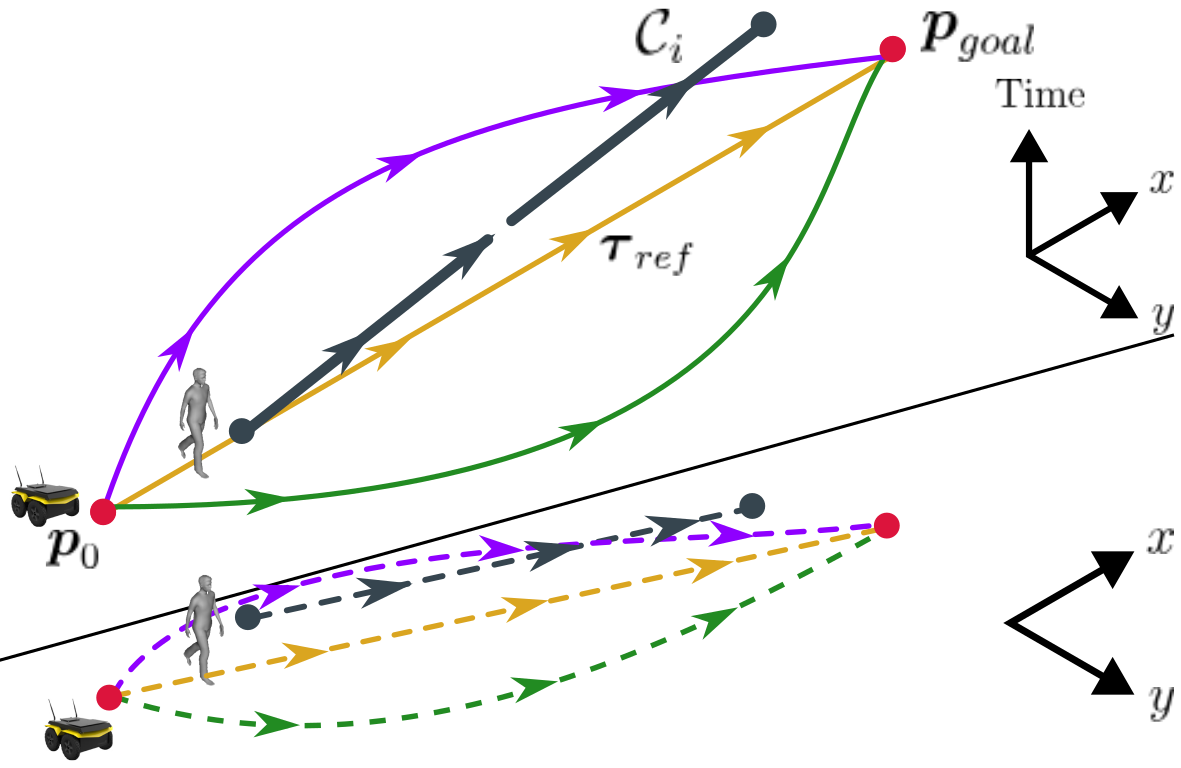}    \\
       \hh{\footnotesize{(b) Trajectories in different homotopy classes and $\boldsymbol{\tau}_{ref}$}}  \\
    \end{tabular}
    \caption{\hh{Schematic representation of trajectories starting at point $\boldsymbol{p}_0$ and ending at $\boldsymbol{p}_{goal}$. Obstacles trajectory is $\mathcal{C}_i$. Green and violet trajectories have the same homotopy class among them. 3-D (top) with one axis representing time and 2-D (bottom) representations are plotted in both (a) and (b). In (b) we introduce a reference trajectory $\boldsymbol{\tau}_{ref}$ (yellow) to assess the homology class of a single trajectory.}}
    \label{fig:homotopy-classes}
\end{figure}


An example \hh{of trajectories in two different homotopy classes} is depicted in \hh{Fig.~\ref{fig:homotopy-classes}(a)}. 
In this work, we use the almost equivalent concept of \textit{homology} classes, similar to previous works~\citep{bhattacharya2012topological, rosmann2017integrated}. 

\begin{definition}
\citep{bhattacharya2012topological}~(Homologous Trajectories) Two trajectories, $\boldsymbol{\tau}_1:[0,1]\rightarrow\mathcal{X}$ and $\boldsymbol{\tau}_2:[0,1]\rightarrow\mathcal{X}$, connected by the same starting and ending point, are homologous if 
\hh{the loop formed by them ($\boldsymbol{\tau}_1 \sqcup -\boldsymbol{\tau}_2$) forms the boundary of a two-dimensional manifold that is immersed \hh{in $\mathcal{X}$} and does not contain nor intersect with any obstacle. Homologous trajectories belong to the same homology class.}
\end{definition}

\hh{Two homotopic trajectories are always homologous. Even though two homologous trajectories may not be homotopic, 
they are very similar in most of robotic applications. Particularly, in the state space $\mathcal{X}$, two homologous trajectories are always homotopic \citep{bhattacharya2011identification}. In this work, we use the homology classes concept because,} contrary to homotopy classes, \hh{which are hard to compute, }homology classes are comparable in practice through their H-signature~\citep{bhattacharya2012topological}. In the 3-dimensional state space, the H-signature can be computed by assuming that the time dimension progresses linearly (e.g., discrete-time). \hh{To do so, each obstacle $i$ may be defined as a vector field that exhibits rotational symmetry around the time-axis, centered in the 3-dimensional obstacle trajectory $\mathcal{C}_i$, as seen in Fig.~\ref{fig:homotopy-classes}(a). The vector field corresponding to obstacle $i$, $\bm{F}_i$, is constructed as an integral over $\mathcal{C}_i$:
\begin{equation}
    \bm{F}_i(\bm{r})=\int_{\mathcal{C}_i}\frac{d\bm{l}\times(\bm{r}-\bm{l})}{|\bm{r}-\bm{l}|^3}
\end{equation}
\noindent where $d\bm{l}$ is an infinitesimal vector element along $\mathcal{C}_i$, $\bm{l}$ a point in $\mathcal{C}_i$ and $\bm{r}$ the point in which the field is being evaluated. This definition is analogous to the Biot-Savart law of electromagnetism.}



\hh{An important property of $\bm{F}_i$ is that it is not path-independent, i.e., the integral of $\bm{F}_i$ between two points depends on the specific path taken, not only on the end-points. Nevertheless, the integral over $\bm{F}_i$ of trajectories that belong to the same homology class is the same. In Fig.~\ref{fig:homotopy-classes}(a), all violet trajectories and all green trajectories exhibit this property, respectively. Thus, path independence holds within homology classes. This is used to define signatures that represent homology classes, named H-signatures. For every trajectory $\boldsymbol{\tau}$, having $M$ obstacles in the environment, its H-signature $\mathcal{H}_3:\mathcal{X}_{traj}\rightarrow \mathbb{R}^M$ is a vector in $\mathbb{R}^M$}:
\begin{equation}
    \mathcal{H}_3(\tau)=[h_1(\boldsymbol{\tau}), h_2(\boldsymbol{\tau}), \dots, h_M(\boldsymbol{\tau})]^T,
    \label{eq:H3}
\end{equation}
where $h_i(\boldsymbol{\tau})$ is defined by the integral:
\hh{\begin{equation}
    h_i(\boldsymbol{\tau}) = \int_{\boldsymbol{\tau}} \bm{F}_i(\bm{r})d\bm{r}.
    \label{eq:signature}
\end{equation}}

\section{Homology class prediction} \label{sec:pred}

In this section, we formally define our version of H-signature and present our approach to predict the homology class selected by humans in crowded scenarios.

\subsection{Modified H-signature} \label{sec:signature} 

\hh{Equation~\ref{eq:signature} holds an important property. The integral of $\bm{F}_i$ over a closed loop $\gamma$, $\oint_\gamma\bm{F}_i(\bm{r})d\bm{r}$, corresponds to the topological winding number of the loop around $\mathcal{C}_i$, i.e., how many times $\gamma$ loops around $\mathcal{C}_i$. For example, if $\oint_\gamma\bm{F}_i(\bm{r})d\bm{r} = 0$, then $\gamma$ does not enclose the obstacle; if $\oint_\gamma\bm{F}_i(\bm{r}) = \pm 1$, it encloses it, with the sign depending on the loop direction; and if $\oint_\gamma\bm{F}_i(\bm{r})d\bm{r} = \pm n_w$, it loops around it $n_w$ times. This property may be used to compare the homology classes of two trajectories, $\boldsymbol{\tau}_1$ and $\boldsymbol{\tau}_2$, connected to the same starting and ending points, by using the loop formed by $\boldsymbol{\tau}_1\sqcup -\boldsymbol{\tau}_2$.:
\begin{equation}
    h_i(\boldsymbol{\tau}_1\sqcup -\boldsymbol{\tau}_2) = \oint_{\boldsymbol{\tau}_1\sqcup -\boldsymbol{\tau}_2} \bm{F}_i(\bm{r})d\bm{r}
    \label{eq:sig-i}
\end{equation}}

Therefore, $h_i(\boldsymbol{\tau}_1\sqcup -\boldsymbol{\tau}_2)$ will be 0 if the trajectories do not enclose the obstacle $i$, and $\pm 1$ if they do. 

We define our own version of the H-signature using a reference trajectory $\boldsymbol{\tau}_{ref}$, represented in \hh{Fig.~\ref{fig:homotopy-classes}(b)} in yellow, and Equation~\ref{eq:sig-i}. We use a reference trajectory to be able to assess the topology of a trajectory in isolation, consistently and invariantly. Having a trajectory $\boldsymbol{\tau}$ with initial and final times $t_0$ and $t_f$, whose origin is $\boldsymbol{\tau}^0 = (x_o, y_o, t_0)$ and final point is $\boldsymbol{\tau}^f = (x_f, y_f, t_f)$; the reference trajectory is defined by the straight line that connects the points $\boldsymbol{\tau}^0$ and $\boldsymbol{\tau}^f$.

Among the infinite set of trajectories, that could be chosen, we selected $\boldsymbol{\tau}_{ref}$ as the reference path for the signature as it is \hh{the simplest trajectory that drives the robot to the goal if there were not obstacles (a straight line with constant velocity)}, and it forms a loop with $\boldsymbol{\tau}$.  
We define the new signature for $M$ obstacles as:
\begin{equation}
    \mathcal{H}_{ref}(\boldsymbol{\tau})=[h_1(\boldsymbol{\tau}_{ref}\sqcup -\boldsymbol{\tau}), \dots, h_M(\boldsymbol{\tau}_{ref}\sqcup -\boldsymbol{\tau})]^T
\end{equation}

 \hh{In practice, and due to our application of $\mathcal{H}_{ref}$, we use the absolute value of the integral as the H-signature, as the sign provides redundant information. In this way, for each obstacle $i$, the element $i$ of $\mathcal{H}_{ref}(\boldsymbol{\tau})$ will have the value 0 if $\boldsymbol{\tau}$ has the same topological behavior as $\boldsymbol{\tau}_{ref}$ with respect to $i$ and $n_w$ otherwise (usually 1 in our application, as robot trajectories are normally not supposed to loop around pedestrians). Unlike $\mathcal{H}_3$ defined in Equation~\ref{eq:H3}, the meaning of $\mathcal{H}_{ref}$ may be easily understood by humans. Intuitively, $\mathcal{H}_3$ encodes how much a trajectory loops around each of the obstacles, with positive or negative values depending on the looping side. Thus, the value of $\mathcal{H}_3(\tau)$ depends on the starting and final position of $\tau$ and the specific trajectories of the obstacles, whereas the value of $\mathcal{H}_{ref}$ may be intuitively inferred. For example, $\mathcal{H}_{ref}$ of the violet trajectory in Fig.~\ref{fig:homotopy-classes}(b) is [1] and $\mathcal{H}_{ref}$ of the green one is [0]. Therefore, } the benefits of the signature are that it is humanly understandable and invariant of the shape and direction of the trajectories\hh{, as well as to the scenarios and environments.} 
 
\begin{figure}
    \centering
    \begin{tabular}{@{}cc@{}}
             \includegraphics[height=5cm]{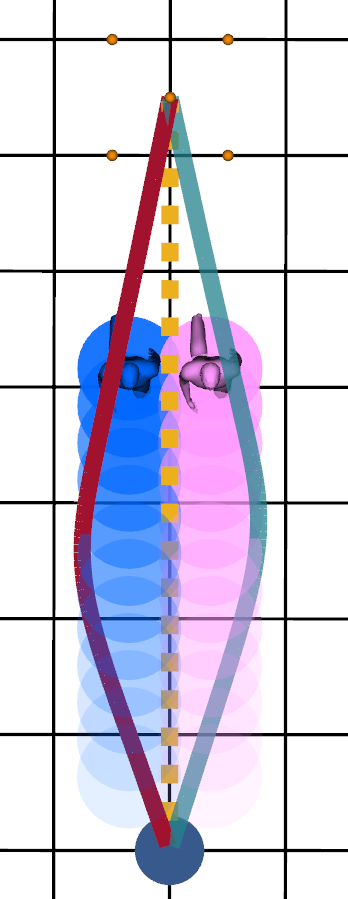} &
         \includegraphics[height=5cm]{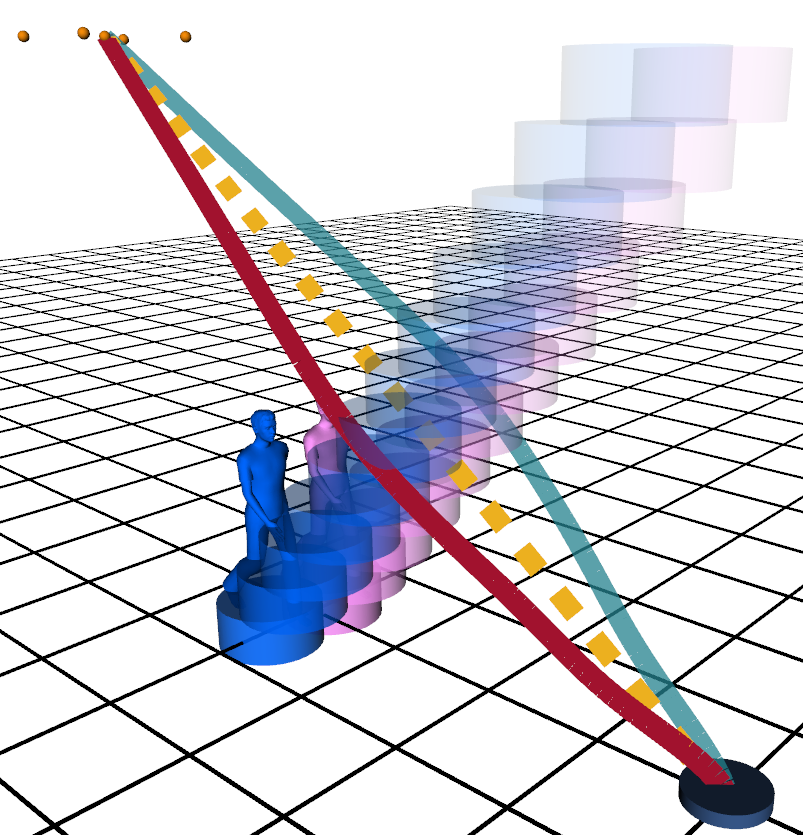}\\
         \footnotesize{(a) Top view} & 
        \footnotesize{(b) State space (time in the z-axis)}
    \end{tabular}
    \caption{An example of two trajectories, left-passing (red line) and right-passing (blue line), in distinct homology classes. The trajectories of the robot (blue disk) pass the future motion prediction of the humans (transparent cylinders) differently. \hh{The reference trajectory $\boldsymbol{\tau}_{ref}$ (dotted yellow) would lead to collision. Orange dots are other possible goal grid points computed by enlarging the goal with the robot radius.}}
    \label{fig:h-ref}
\end{figure}

 \hh{Every homology class may not have non-colliding and feasible trajectories when considering robot and human radius, specially in crowded environments. For example, Fig.~\ref{fig:h-ref} shows a scenario with two humans where the robot may only choose between trajectories in two homology classes. Even though $\boldsymbol{\tau}_{ref}$ leads to a collision, $\mathcal{H}_{ref}$ may still be computed.} 

\subsection{Homology class selection}

While humans are constantly choosing a suitable topological path when navigating, manually defining a function that considers all factors that are involved in the decision-making process is impossible. We aim to learn the human selection process so that, having a recorded real-world scenario where a human has followed the ground truth trajectory $\boldsymbol{\tau}^h$, our method is able to predict $\mathcal{H}_{ref}(\boldsymbol{\tau}^h)$ without knowing $\boldsymbol{\tau}^h$. This knowledge could be applied to plan a global guidance trajectory $\boldsymbol{\tau}^g$ that belongs to the same homology class as $\boldsymbol{\tau}^h$ and use a local planner to navigate within that class to achieve socially compliant navigation. However, sampling a guidance trajectory having a H-signature is a complex task that is sometimes unsolvable, since there may be H-signatures where all trajectories lead to collisions. Therefore, instead of simply trying to predict $\mathcal{H}_{ref}(\boldsymbol{\tau}^h)$ in a scenario, we propose sampling a guidance trajectory $\boldsymbol{\tau}^g$ and estimating the difference between $\mathcal{H}_{ref}(\boldsymbol{\tau}^g)$ and $\mathcal{H}_{ref}(\boldsymbol{\tau}^h)$. The process is repeated for guidance trajectories that belong to every possible homology class in the environment, so that the one with the lowest difference is estimated to be the most social one. 

\begin{figure*}
    \centering
    \includegraphics[width=0.95\textwidth]{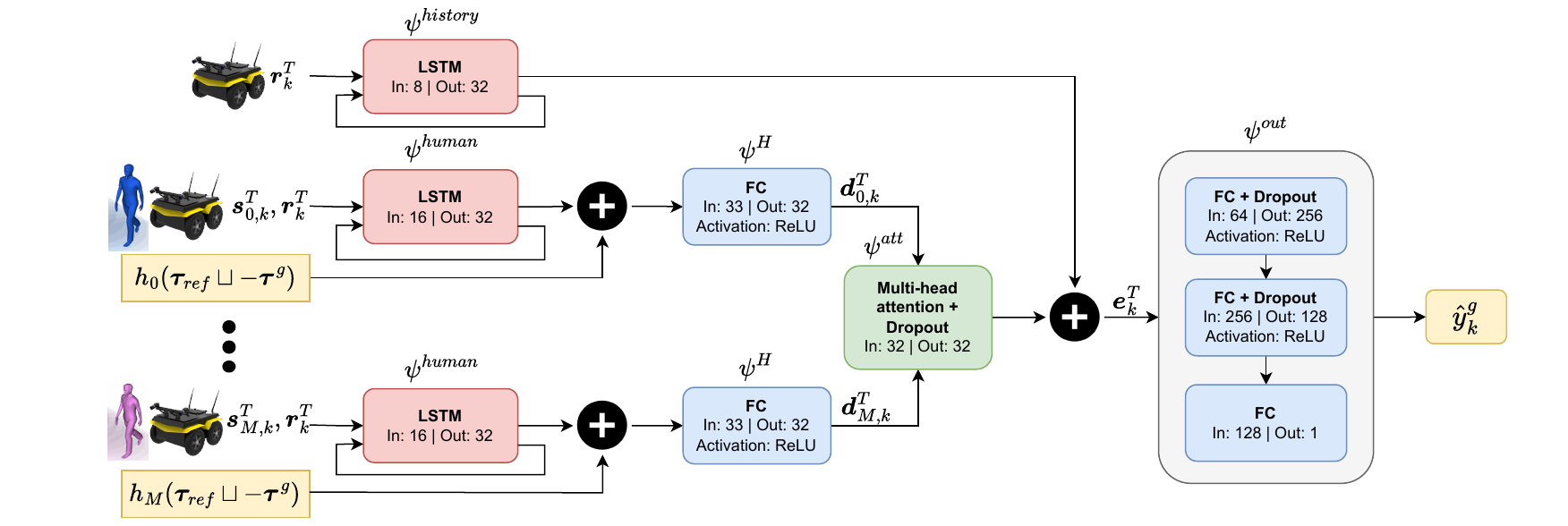}
    \caption{Diagram with the layers of the network that we use to infer social high-level decisions. The inputs are the historical state of the robot, $\boldsymbol{r}^T_k$, and the historical state of the $M$ surrounding humans, $\boldsymbol{s}^T_k$, at time $k$ over the last time horizon $T$ and the H-signature of the proposed guidance, $\mathcal{H}_{ref}(\boldsymbol{\tau}^g)$. The output, $\hat{y}^g_k$, is the estimation of the result of Equation~\ref{eq:MSE}, which is the estimated difference of the homology class of guidance and the preference of a human.}
    \label{fig:network}
\end{figure*}

We consider a parametric cost function that estimates that difference using as inputs previous observations of the robot's positions $\boldsymbol{p}_k^T$ and the other humans' positions in the scenario $\mathcal{O}^T_k$ over a time horizon $T$ at time $k$. 
The cost function designed $\mathcal{J}_\theta(\mathcal{H}_{ref}(\boldsymbol{\tau}^g), \boldsymbol{p}_k^T, \mathcal{O}^T_k)$ is approximated using a deep neural network \hh{with parameters $\theta$}, represented in a diagram in Fig.~\ref{fig:network}. The goal of the network is to estimate the difference in H-signature (in a Euclidean sense) between the trajectory $\boldsymbol{\tau}^g$ and the trajectory $\boldsymbol{\tau}^h$ that a human would choose: 

\begin{equation}
    y^g = MSE(\mathcal{H}_{ref}(\boldsymbol{\tau}^g), \mathcal{H}_{ref}(\boldsymbol{\tau}^h)),
    \label{eq:MSE}
\end{equation}

\noindent where $y^g$ is the desired output of the network. 

The network structure follows the architecture proposed in \citet{SalzmannIvanovicEtAl2020} and \citet{IvanovicHarrisonEtAl2023}. It shares the same encoding architecture with modifications to include the homology information, and fully connected layers to finally output the desired cost. The states of the robot and the surrounding humans are first extended by linearly interpolating the velocity ($v_{x,k}, v_{y,k}$), acceleration ($a_{x,k}, a_{y,k}$), and the heading angle $\alpha$ from current and previous positions. Then, the considered full state of the robot at time $k$ is $\boldsymbol{r}_k=[x_k,y_k,v_{x,k},v_{y,k},a_{x,k},a_{y,k},\sin{\alpha_k},\cos{\alpha_k}]$ and the state of the human $j$ is $\boldsymbol{s}_{j,k}=[x_{j,k},y_{j,k},v_{j,x,k},v_{j,y,k},a_{j,x,k},a_{j,y,k},\sin{\alpha_{j,k}},\cos{\alpha_{j,k}}]$. The full state of the robot and the human $j$ over the last time horizon $T$ are denoted as $\boldsymbol{r}^T_k = [\boldsymbol{r}_{k-T}, \dots, \boldsymbol{r}_k]$ and $\boldsymbol{s}^{T}_{j,k} = [\boldsymbol{s}_{j,k-T}, \dots, \boldsymbol{s}_{j,k}]$, respectively, and the full state of all the obstacles in the environment is denoted as $\mathcal{S}^T_k=[\boldsymbol{s}^{T}_{0,k}, \dots, \boldsymbol{s}^{T}_{M,k}]$, given that there are $M$ humans influencing the robot at time $k$.

The state of the robot over the time horizon, $\boldsymbol{r}^T_k$, is encoded using an LSTM layer, $\psi^{history}(\boldsymbol{r}^T_k)$, so that the previous states are considered while keeping time dependencies. The same approach is taken to encode the surrounding humans' state but, in this case, the state of each of the human $j$ at each time step $k$, $\boldsymbol{s}_{j,k}$, is concatenated to the state of the robot at the same time step, $\boldsymbol{r}_k$, and given as input to an LSTM encoder, using the same weights for every human: $\psi^{human}(\boldsymbol{s}^T_{j,k}, \boldsymbol{r}^T_k)$. The output of human $j$ is concatenated to its corresponding value of $\mathcal{H}_{ref}$ signature, $h_j(\boldsymbol{\tau}_{ref}\sqcup -\boldsymbol{\tau}^g)$. This is fed into a Fully Connected Layer (FCL), $\psi^{H}$, to get an embedding vector, $\boldsymbol{d}^T_{j,k}$, that encodes the information of the human $j$ and its relation with $\boldsymbol{\tau}^g$:

\begin{equation}
    \boldsymbol{d}^T_{j,k} = \psi^{H}(\psi^{human}(\boldsymbol{s}^T_{j,k}, \boldsymbol{r}^T_k), h_j(\boldsymbol{\tau}_{ref}\sqcup -\boldsymbol{\tau}^g))
\end{equation}

The output of the $\psi^{H}$ layers is aggregated using attention \citep{vaswani2017attention} ($\psi^{att}$), as in \citet{vemula2018social}, to combine all the influence that the neighbor obstacles have in the robot into a single vector, which is concatenated to the robot encoded state to get the final representation state, $\boldsymbol{e}^T_k$:
\begin{equation}
    \boldsymbol{e}^T_k = \{\psi^{history}(\boldsymbol{r}^T_k), \psi^{att}(\boldsymbol{d}^T_{0,k}, \dots, \boldsymbol{d}^T_{M,k})\}
\end{equation}

Finally, $\boldsymbol{e}^T_k$ is fed into a model with 3 fully connected layers, $\psi^{out}$. The two intermediate models, as well as $\psi^{H}$, include ReLU activation functions and dropout layers. The output is a single value, which is the cost of choosing the homology class:

\begin{equation}
    \mathcal{J}_\theta(\mathcal{H}_{ref}(\boldsymbol{\tau}^g), \boldsymbol{p}_k^T, \mathcal{O}^T_k) = \hat{y}^g_k = \psi^{out}(\boldsymbol{e}^T_k)
    \label{eq:net-cost}
\end{equation}

\subsection{Training on real-world data}

Real-world datasets of humans navigating in crowds are used to train the network, particularly the UCY \citep{lerner2007crowds} and ETH \citep{pellegrini2009you} datasets. The training tries to extract the behavior of every of the humans of the dataset. Because the data is prerecorded, the history and future motion of the humans are known. Each training sample consists of a planning problem for one of the humans where the future is used as ground truth. The network parameters $\theta$ are optimized to fit the data:

\begin{equation}
    \min_\theta\sum_{k=1}^{N_s}\sum_{g=1}^{N_h} \mathcal{L}_{pred}(\hat{y}^g_{k}, y^
    g_{k}),
\end{equation}
\noindent where $\mathcal{L}_{pred}$ is chosen to be Mean Squared Error, $N_s$ are the number of scenarios in the dataset, $N_h$ the number of homology classes found in scenario $k$, $\boldsymbol{\tau}^g$ a guidance trajectory of a homology class, $\hat{y}^g_{k}$ the output of the network for the guidance $\boldsymbol{\tau}^g$ in scenario $k$ and $y^g_{k}$ the difference between the H-signature of the guidance $\boldsymbol{\tau}_g$ and the actual trajectory followed by the human, as in Equation~\ref{eq:MSE}. The high-level training algorithm is represented in Algorithm~\ref{alg:training}.

\begin{algorithm}
\caption{Training}\label{alg:training}
\SetKwInput{KwInput}{Input}                
\SetKwInput{KwOutput}{Output}              
\DontPrintSemicolon
  
\KwInput{Dataset of pedestrians navigating in a crowd in time $\mathcal{D}$}
\KwOutput{Updated network weights $\theta$}
\For{$training\_iterations$}{
$\boldsymbol{p}_k^T, \mathcal{O}^T_k, \boldsymbol{\tau}^h_{k}, \mathcal{O}^{fut}_k \gets SampleBatch()$\;
\tcc{Sample batch with size $B_s$ of different scenarios with previous and future agent and obstacles positions over a time horizon. $k\in(0,\dots,B_s)$ is one element of the batch.}
$\mathcal{T}^*_k \gets SampleGuidances(\mathcal{O}^{fut}_k, \boldsymbol{p}_k, \boldsymbol{p}_{goal})$\;
\tcc{Sample up to $N$ topologically distinct guidance trajectories for all the $B_s$ scenarios ($B_s$x$N$ guidances).}
$\boldsymbol{y}_k\gets MSE(\mathcal{H}_{ref}(\mathcal{T}^*_k), \mathcal{H}_{ref}(\boldsymbol{\tau}^h_k))$\;
\tcc{Compute the actual homological difference between the human's trajectory and each of the guidances ($B_s$x$N$).}
$\boldsymbol{\hat{y}}_k \gets \mathcal{J}_\theta(\mathcal{H}_{ref}(\mathcal{T}^*_k), \boldsymbol{p}_k^T, \mathcal{O}^T_k)$\;
\tcc{Estimate the cost using the network (Real batch size is $B_s$x$N$).}
$\theta \gets UpdateWeights(\theta, \mathcal{L}_{pred}(\boldsymbol{\hat{y}}_k, \boldsymbol{y}_k))$\;
}
\end{algorithm}

The network is designed so that the different homology classes are loaded as different entries in the same batch. Therefore, the inference and training is done in parallel for all the homology classes, and there is no extra computational cost of having more or less homology classes. 

\textbf{Accounting for scalability.} Our approach exhibits effective scaling in three different ways. First, it may have inputs with different numbers of previous timesteps, such as humans that were not previously sensed (their trajectory is shorter). This is achieved by using LSTM layers to process the temporal data used for the input of the network. Second, the number of surrounding humans is not fixed and does not need to be always the same, as the attention layer deals with it. Finally, it is scalable in the lengths and shapes of the trajectories, as the H-signature depends on the obstacles instead of the specific guidance trajectory. We show this in Section~\ref{sec:pred-eva} by testing the method with a prediction horizon much higher than the one used in training. In addition, the network could be easily extended to include other features such as the map of the environment as in the Trajectron++~\citep{SalzmannIvanovicEtAl2020} original work.

\textbf{Accounting for multi-modality.} Human navigation is multi-modal. In our approach, each of the navigation modes are represented with homology classes. A cost is assigned to each of the homology classes and the one with the minimum cost is chosen, as it is the one estimated to be the most likely to be followed by a human. As the cost is computed for all the homologies, the ones with similar cost 
are similarly acceptable. Nonetheless, in our navigation approach, the robot motion should be predictable, so the most probable mode is always chosen (the one with the minimum cost).

\section{Navigation system} \label{sec:system}

\begin{figure*}
    \centering
    \includegraphics[width=0.99\textwidth]{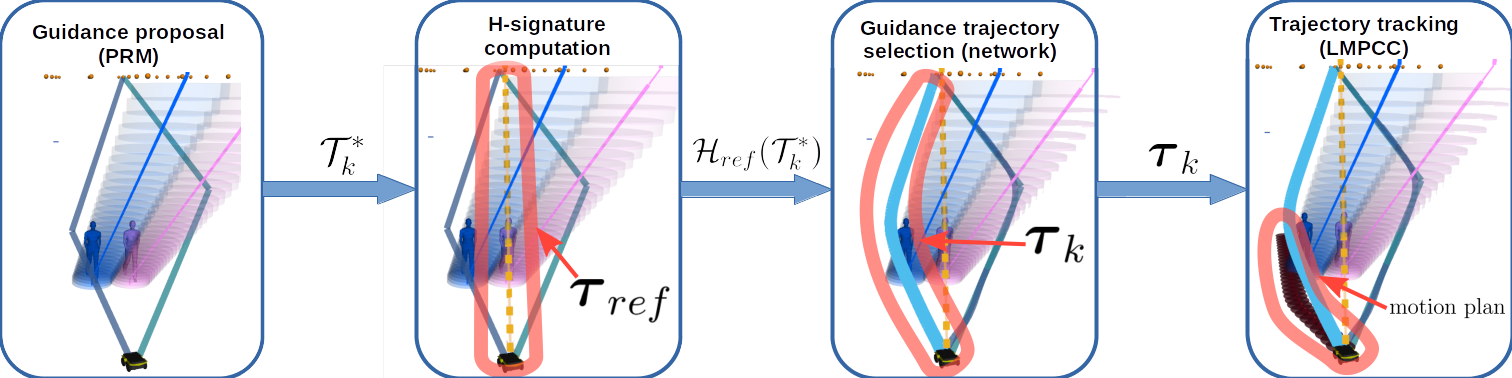}
    \caption{\hh{Schematic representation of SHINE. First, the guidance proposal algorithm in Section~\ref{sec:PRM} finds trajectories (two blue lines originating from the robot) in different homology classes ($\mathcal{T}^*_k$). Second, each trajectory is compared to a reference trajectory to identify its homology class. Third, the guidance selection component in Section~\ref{sec:pred}, that was trained to select socially compliant trajectories from human motion data, selects the guidance trajectory ($\tau_k$). Finally, the MPC-based planner in Section~\ref{sec:LMPCC} computes the motion commands based on $\tau_k$.}}
    \label{fig:components}
\end{figure*}

In the following, we present a navigation framework which derives topologically distinct trajectories, selects one guidance trajectory using the learned cost function, and applies a local motion planner to track it. A schematic representation of the three main components of the system is provided in Fig.~\ref{fig:components}.

\subsection{Guidance trajectories proposal} \label{sec:PRM}

We use the algorithm proposed in \citet{de2023globally} to compute a sparse representation of paths (guidance trajectories) that go from the initial position of the robot to the goal, and use the H-signature to filter the trajectories 
that 
belong to 
the same homology class. 

The method is built on Visibility-PRM and works in the same state space introduced in the problem formulation, composed by the workspace and time $\mathcal{X}=\mathbb{R}^2\times[0,T]$. The idea is to keep a graph of guard and connector nodes initialized with the initial and final position of the desired trajectory as guard nodes. New nodes are randomly sampled and added to the graph if they can be connected without colliding with static and dynamic obstacles. 
If a new node connects two guards, it is added as a connector. Otherwise, it is included as a guard. If two connectors connect the same two guards and the paths they construct are homologous, only the connector with the shortest path is kept. Therefore, the result of the algorithm is a graph with a set of trajectories $\mathcal{T}^*$ where $\mathcal{H}_{ref}(\boldsymbol{\tau}_1)\neq\mathcal{H}_{ref}(\boldsymbol{\tau}_2), \forall (\boldsymbol{\tau}_1, \boldsymbol{\tau}_2)\in\mathcal{T}^*$. Each of the resulting trajectories is topologically distinct and represents a homology class that the robot may follow to reach the goal.

The minimum-cost guidance trajectory, having the cost of Equation~\ref{eq:net-cost}, can be selected as:
\begin{equation}
    \boldsymbol{\tau}_k = \underset{\boldsymbol{\tau}_{i,k} \in \mathcal{T}_k^*}{\arg\min} \mathcal{J}_\theta(\mathcal{H}_{ref}(\boldsymbol{\tau}_{i,k}), \boldsymbol{p}_k^T, \mathcal{O}^T_k)
\end{equation}

\subsection{Local Optimization-based Planner}\label{sec:LMPCC}
The selected guidance trajectory identifies the most suitable homology class but is not guaranteed to be dynamically feasible or collision free. To obtain a high-quality trajectory, we initialize and track the guidance trajectory with a local optimization-based planner that refines the trajectory in the same state space, similar to~\citep{de2023globally}. The local planner is an MPC that solves a trajectory optimization \hh{problem}, enforcing dynamic and collision avoidance constraints and returns a smooth trajectory near the guidance trajectory. We use the planner LMPCC~\citep{brito2019model}. From the guidance trajectory, we first derive a guidance \textit{path}, $[0, S]\to\mathbb{R}^2$ that maps the distance along the path (of length $S$) to an $x,y$-position. The MPC solves the optimization problem
\begin{subequations}\label{eq:mpc_optimization}
\begin{align}
    \min_{\b{u} \in \mathbb{R}^{n_u}, \b{x}\in\mathbb{R}^{n_x}} \hspace{0.5em} & \sum_{k = 0}^N J_k \label{eq:mpc_objective}\\
    \textrm{s.t.} \hspace{1.5em} & \b{x}_0 = \b{x}_{\textrm{init}} \label{eq:mpc_init}\\
& \b{x}_{k + 1} = f(\b{x}_k, \b{u}_k), \ k = 0, \hdots, N-1 \label{eq:mpc_dynamics}\\
  &g(\b{x}_k, \b{o}_{j,k})\leq 0,\ k = 1, \hdots, N .\label{eq:mpc_constraint}%
\end{align}
\end{subequations}
The cost function at step $k$, with weights $w$, is given by
\begin{equation}
    J_k = w_cJ_{c, k} + w_lJ_{l, k} + w_vJ_{v, k} + w_aJ_{a, k} + w_{\omega}J_{\omega, k},
\end{equation}
where $J_{c, k}$ and $J_{l, k}$ are contour and lag costs that track the guidance path as defined in~\citep{brito2019model}, $J_{v, k} = ||v_k - v^{\textrm{guidance}}_k||_2^2$ tracks the velocity along the guidance trajectory and, $J_a = ||a||_2^2$ and $J_{\omega} = ||\omega||_2^2$ weigh the control inputs. The constraints in Eqs.~\eqref{eq:mpc_init} and~\eqref{eq:mpc_dynamics} constrain the initial state and dynamics, respectively. Collision avoidance is specified through the constraints $g(\b{x}_k, \b{o}_{j,k})\leq 0$,
\begin{equation}
        g(\b{x}_k, \b{o}_{j,k}) = 1-(\Delta\b{p}_{j,k})^T\b{R}(\phi)^T\begin{bmatrix}\frac{1}{r^2} & 0 \\ 0 & \frac{1}{r^2}\end{bmatrix}\b{R}(\phi)(\Delta\b{p}_{j,k}), \label{eq:deterministic-collision-constraints}
\end{equation}
where $\Delta\b{p}_{j,k} = \b{p}_k - \b{o}_{j,k}$, $\b{R}(\phi)$ is a rotation matrix with orientation $\phi$ of the robot and $r = r_{{robot}} + r_{{obs}}$. The constraints specify that the robot's and obstacles' discs should not overlap. The robot dynamics are modeled as a second-order unicycle model~\citep{siegwart_introduction_2011}. We solve the online optimization with Forces Pro~\citep{domahidi_forces_2014}.


\subsection{Enforcing consistent decisions}

At every time step $k$, new guidance trajectories are sampled and a new homology class is selected. In this way, our approach may react to changes in environments with fast replanning capabilities, accounting for estimation errors or changes in the behavior of the pedestrians. Nevertheless, fast switching between trajectories in different topology classes can lead to indecisive and ultimately non-social navigation.

The problem is partially solved by the network, as people's intentions are usually constant in time and that behavior is intrinsically learned. In addition, we include a consistency weight $w_c \in [0,1)$ to encourage the robot to keep navigating within the same homology class. The weight is multiplied by the estimated network cost only if the homology class of a trajectory guidance is the same as the homology class followed in the previous control period, reducing its selection cost. 

The high-level navigation algorithm is represented in Algorithm~\ref{alg:navigation}, where the consistency weight is applied in line~\ref{line:wc}.

\begin{algorithm}
\caption{Navigation}\label{alg:navigation}
\SetKwInput{KwInput}{Input}                
\SetKwInput{KwOutput}{Output}              
\DontPrintSemicolon
  
\KwInput{Previous robot positions $\boldsymbol{p}_k^T$, robot goal $\boldsymbol{p}_{goal}$, previous obstacle position $\mathcal{O}^T_k$, and predictions of future positions of the obstacles $\mathcal{O}^{fut}_k$}
\KwData{Consistency weight $w_c$}
\For{Every control period}{
$\mathcal{T}^*_k \gets SampleGuidances(\mathcal{O}^{fut}_k, \boldsymbol{p}_k, \boldsymbol{p}_{goal})$\;
$\boldsymbol{\hat{y}}_k \gets \mathcal{J}_\theta(\mathcal{H}_{ref}(\mathcal{T}^*_k), \boldsymbol{p}_k^T, \mathcal{O}^T_k)$\;
\tcc{Estimate the cost of every guidance using the network.}
\For{$\boldsymbol{\tau}_{i,k} \in \mathcal{T}^*_k$}{
\If{$\mathcal{H}_{ref}(\boldsymbol{\tau}_{i,k})=\mathcal{H}_{ref}(\boldsymbol{\tau}_{k-1})$}
{
$\hat{y}^i_k \gets \hat{y}^i_k * w_c$\; \label{line:wc}
\tcc{Apply consistency weight.}
}
}
$\boldsymbol{\tau}_{k} \gets argmin(\mathcal{T}^*_k, \boldsymbol{\hat{y}}_k)$\;
\tcc{Select guidance with the minimum estimated cost.}
$\boldsymbol{u}_{k} \gets LMPCC(\boldsymbol{\tau}_{k})$\;
\tcc{Use LMPCC to get the control commands.}
$ApplyControl(\boldsymbol{u}_{k})$
}
\end{algorithm}
\section{Evaluation} \label{sec:exp}

This section discusses the experimental setup and results. We explain the experiments we conducted using real-world datasets in Section~\ref{sec:pred-eva}, in simulation in Section~\ref{sec:simu-eval}, and in a real platform in Section~\ref{sec:real-eva}.

\subsection{Experimental setup} \label{sec:experimental-setup}

\textbf{Training setup.} The LSTM and attention layer parameters are initialized with the pretrained weights of Trajectron++~\citep{IvanovicHarrisonEtAl2023} provided in their original work, to leverage their feature extraction. Some of the most important parameters of the network training are indicated in Table~\ref{tab:network-param}. We included L2 regularization. We used a history time horizon of 2.8 s and a prediction horizon of 4.8 s, as they are commonly employed in the ETH/UCY dataset.

\begin{table}
    \centering
    \caption{Training parameters used in the network.}
    \begin{tabular}{|c|c|}
        \hline
        \textbf{Parameter} & \textbf{Value} \\
        \hline
         Batch size & 32 \\
         Training epochs & 10 \\
         Learning rate & 0.0001 \\
         Optimizer & Adam \\
         History time horizon & 2.8 s \\ 
         Prediction time horizon & 4.8 s \\ 
         Dropout probability & 0.1 \\
         \hline
    \end{tabular}
    
    \label{tab:network-param}
\end{table}

\textbf{Hardware setup.} The real-world evaluation is conducted on a Clearpath Jackal mobile robot in a room of size 7.5 m x 6.5 m. The platform is equipped with an Intel i5 CPU@2.6GHz. The robot and pedestrian positions are determined through a marker-based tracking system. Pedestrian predictions are based on the assumption of constant velocity, with the velocity obtained using a Kalman filter. 

The motion planner is implemented in C++/ROS and the homology selector in Python. The whole system will be released as open source.

\subsection{Prediction evaluation} \label{sec:pred-eva}

The capacity of the network to select the correct guidance trajectory is evaluated in these experiments. To do so, we use testing scenarios of the ETH/UCY dataset where more than one homology class could be selected to navigate. We designed an accuracy score that measures the number of times the homology class actually chosen by the human is the one with the lowest cost. In the ETH/UCY dataset, there are five different environments: ETH, Hotel, University, Zara1 and Zara2. To evaluate the network and its generalization capabilities, it was trained in four scenarios and evaluated in the other one (Leave One Out). 

As baselines, we propose selecting the guidance trajectory using three hand-crafted costs based on heuristics, taking $N_\tau$ samples of positions $\boldsymbol{p}_i$ and accelerations $\boldsymbol{a}_i$ at constant time intervals along the trajectories:

\begin{itemize}
    \item Minimum length: The shortest guidance trajectory has the lowest cost: $\mathcal{J}_{len} = \sum_{i \in N_\tau} ||\boldsymbol{p}_i-\boldsymbol{p}_{i-1}||$
    \item Minimum acceleration: The smoothest trajectory has the minimum cost: $\mathcal{J}_{acc} = \sum_{i \in N_\tau} \alpha^i||\boldsymbol{a}_i||$, where $\alpha\approx 1$ to discount accelerations in time \hh{(0.95 in the implementation)}.
    \item Mixed cost: Combine previous costs: $\mathcal{J}_{mix} = \mathcal{J}_{len} + \mathcal{J}_{acc}$
\end{itemize} 

Additionally, we tried an alternative H-signature, $\mathcal{H}_{right}$ to support the use of our selected H-signature, $\mathcal{H}_{ref}$. We define the new H-signature in the same way as in Section~\ref{sec:signature} with a reference trajectory that surrounds the whole environment through the right side, $\boldsymbol{\tau}_{right}$. Having a trajectory $\boldsymbol{\tau}$ with initial and final times $t_0$ and $t_f$, whose origin is $(x_o, y_o, t_0)$ and final point is $(x_f, y_f, t_f)$; and a set of positions of obstacles in that time interval $\mathcal{O}^{t_f-t_0}_{t_f}$ where the set of coordinates in the y-axis of the all the positions are $\boldsymbol{y}_\mathcal{O}$; the reference trajectory is defined by the straight lines that connect the following points:

\begin{enumerate}
    \item $\boldsymbol{\tau}^0_{right} = (x_o, y_o, t_0)$
    \item $\boldsymbol{\tau}^1_{right} = (x_o, y_o, t'_0), \enskip t'_0 < t_0$
    \item $\boldsymbol{\tau}^2_{right} = (x_o, y_{right}, t'_0), \enskip y_{right} < y \enskip \forall y \in \boldsymbol{y}_\mathcal{O}$
    \item $\boldsymbol{\tau}^3_{right} = (x_o, y_{right}, t'_f), \enskip t'_f > t_f$
    \item $\boldsymbol{\tau}^4_{right} = (x_f, y_{right}, t'_f)$
    \item $\boldsymbol{\tau}^5_{right} = (x_f, y_f, t'_f)$    
    \item $\boldsymbol{\tau}^6_{right} = (x_f, y_f, t_f)$
\end{enumerate}

\begin{figure}
    \centering
    \begin{tabular}{@{}cc@{}}
         \includegraphics[height=4cm]{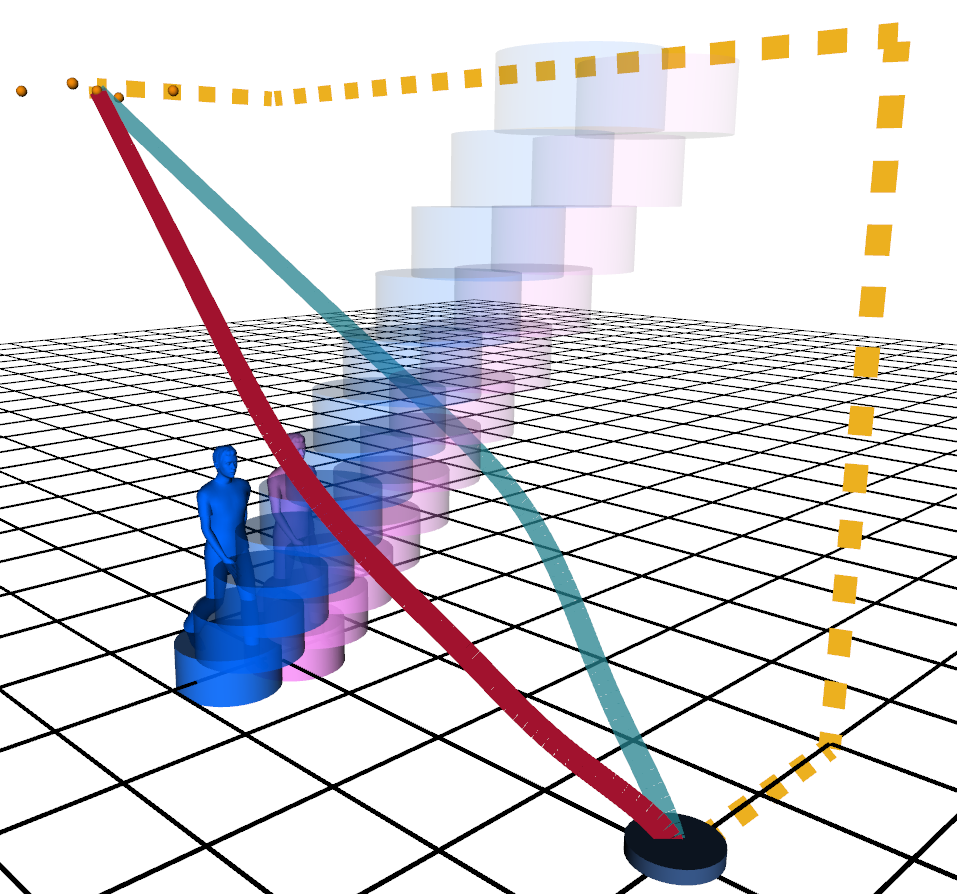}  & 
         \includegraphics[height=4cm]{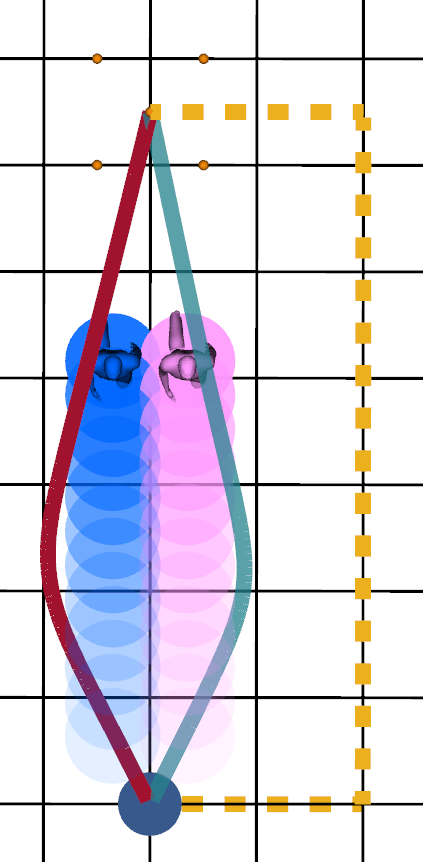}\\
         \footnotesize{(a)} & 
        \footnotesize{(b)}
    \end{tabular}
    \caption{Two views of $\boldsymbol{\tau}_{right}$ trajectory (yellow) and two trajectories that belong to different homology classes regarding two obstacles, where the axis-z represents time. In (a), the trajectories are represented with the time axis, while (b) is a top view of (a). The $\mathcal{H}_{right}$ of the blue trajectory is [0,0] and the one of the red trajectory is [1,1].}
    \label{fig:h-right}
\end{figure}

The trajectory is represented in Figure~\ref{fig:h-right}, in the same way $\boldsymbol{\tau}_{ref}$ was represented in Figure~\ref{fig:h-ref}. With $M$ humans influencing the robot, we define the new signature as:
\begin{equation}
    \mathcal{H}_{right}(\boldsymbol{\tau})=[h_1(\boldsymbol{\tau}_{right}\sqcup -\boldsymbol{\tau}), \dots, h_M(\boldsymbol{\tau}_{right}\sqcup -\boldsymbol{\tau})]^T,
\end{equation}

For every obstacle $i$, $\mathcal{H}_{right}(\boldsymbol{\tau})$ will be 0 if $\boldsymbol{\tau}$ avoids obstacle $i$ through the right side, and 1 if it avoids it through the left. The network was trained using $\mathcal{H}_{right}$ instead of $\mathcal{H}_{ref}$ to test this other signature. In this section, we refer as $J_{\theta, ref}$ the cost estimated by the network using $\mathcal{H}_{ref}$, defined in Equation~\ref{eq:net-cost}; and $J_{\theta, right}$ the alternative using $\mathcal{H}_{right}$.

The accuracy results are shown in Table~\ref{tab:loo}. Our method significantly outperforms the other approaches. We noticed that, once the human makes a decision on how to pass other humans, it usually sticks with it, following a previously made decision. This explains why all methods achieve good accuracy ($>0.5$). The results thus show that our method does not only select just the homology class the human was previously following, but also accurately estimates the decision at any time step. 


\begin{table}[ht]
    \caption{Accuracy metrics of the predictions of the human homology class.}
    \centering
    \resizebox{\linewidth}{!}{
    \begin{tabular}{|c|ccccc|c|}
        \hline
        \textbf{Cost} & \textbf{ETH} & \textbf{Hotel} & \textbf{Univ} & \textbf{Zara1} & \textbf{Zara2} & \textbf{Avg.} \\
        \hline
         $\mathcal{J}_{len}$   & 0.770 & 0.808 & 0.730 & 0.786 & 0.767 & 0.772 \\
         $\mathcal{J}_{acc}$  & 0.623 & 0.617 & 0.625 & 0.516 & 0.645 & 0.605 \\
         $\mathcal{J}_{mix}$   & 0.730 & 0.797 & 0.729 & 0.779 & 0.758 & 0.759 \\
         $\mathcal{J}_{\theta, right}$ (ours)   & 0.800 & 0.729 & 0.686 & 0.851 & 0.678 & 0.749 \\ 
         $\mathcal{J}_{\theta, ref}$ (ours)       & \textbf{0.984} & \textbf{0.951} & \textbf{0.942} & \textbf{0.955} & \textbf{0.925} & \textbf{0.951} \\
        \hline
    \end{tabular}}
    \label{tab:loo}
\end{table}

\begin{figure}
    \centering
    \begin{tabular}{@{}cc@{}}
         \includegraphics[height=7.5cm]{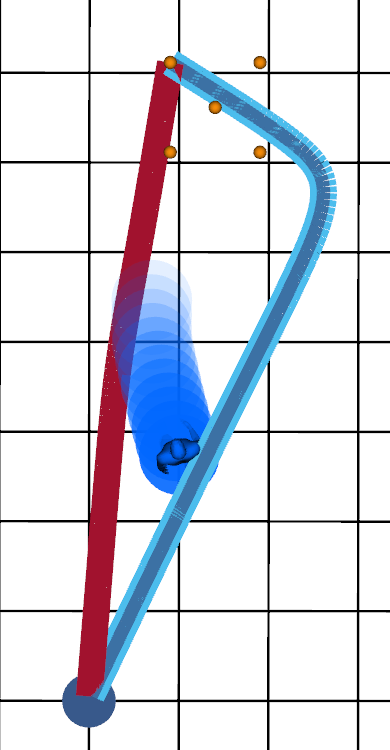}  & 
         \includegraphics[height=7.5cm]{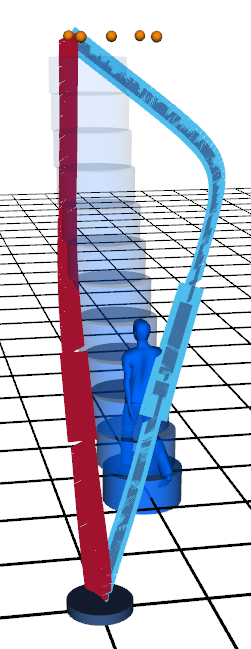}\\
         \footnotesize{(a)} & 
        \footnotesize{(b)}
    \end{tabular}
    \caption{Top (a) and side (b) view of a scenario of the real-world dataset where the guidance chosen by our method (light blue) avoids passing through the pedestrian trajectory and is the one chosen by the real human. The guidance trajectory chosen with $\mathcal{J}_{len}$ is represented in red.}
    \label{fig:not-rude}
\end{figure}

\begin{figure}
    \centering
    \begin{tabular}{@{}cc@{}}
         \includegraphics[height=4cm]{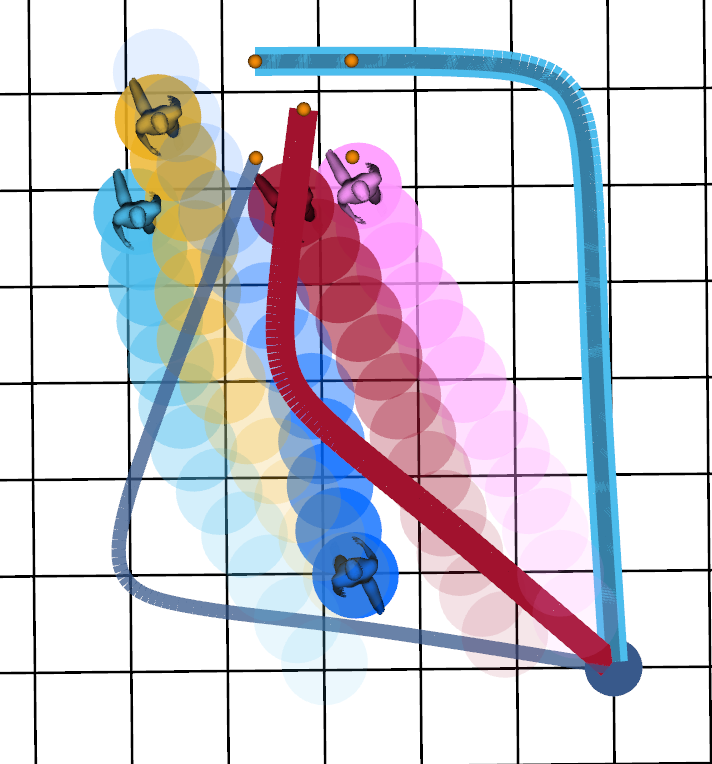}  & 
         \includegraphics[height=4cm]{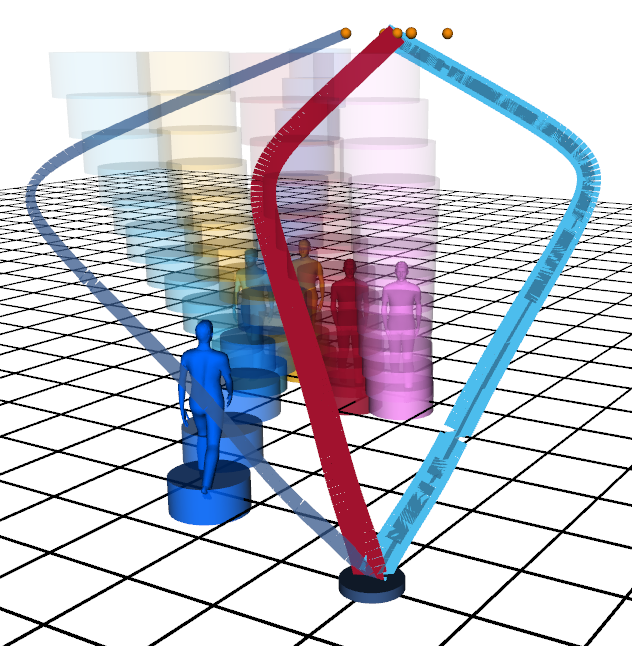}\\
         \footnotesize{(a)} & 
        \footnotesize{(b)}
    \end{tabular}
    \caption{Top (a) and side (b) view of a scenario of the real-world dataset where the guidance chosen by our method (light blue) is less intrusive and is the one chosen by the real human. The guidance trajectory chosen with $\mathcal{J}_{len}$ is represented in red, while in dark blue another guidance trajectory is represented.}
    \label{fig:complex}
\end{figure}

For example, Figure~\ref{fig:not-rude} and Figure~\ref{fig:complex} show scenarios of the dataset where the guidance chosen with $\mathcal{J}_{len}$ cost (red) is different from the one chosen with the network (light blue), and the one chosen with the network corresponds to the actual recorded behavior of the human. In both figures, the future positions of the humans are represented with circles increasing linearly in the z-axis, which represent the time. The guidance trajectories are represented as lines from the initial position to the goal, represented with the area with small orange spheres. In Figure~\ref{fig:not-rude}, the optimal behavior in terms of path length is passing through the left of the blue pedestrian. Nevertheless, this could be socially rude, as it is overtaking the pedestrian from behind while intersecting with its current trajectory, so the network proposes avoiding it through its right side. In Figure~\ref{fig:complex}, the guidance chosen is avoiding every pedestrian through the right, as it is the less invasive option from the three proposed guidance trajectories. 

In addition, it is interesting to see that in Table~\ref{tab:loo} the metrics are consistent among the different unseen environments. This highlights the fact that designing a cost that adapts to different scenarios is hard, while our approach may learn how to navigate in different scenarios. 

The other conclusion extracted from the table is the benefit of using our signature. The input information of the network using $\mathcal{H}_{ref}$ and $\mathcal{H}_{right}$ is the same but with different encoding, but the information of $\mathcal{H}_{ref}$ is easier for the network to understand. The information encoded by $\mathcal{H}_{ref}$ follows the human reasoning. We hypothesize that, when a human tries to reach a goal, it first thinks about the shortest path, and then focus on what obstacles to avoid with respect to that path. 

\textbf{Scalability.} The network faces three different types of scalability in the predictions: In the number of homologies available in the scenario, the number of surrounding humans and the trajectory guidance length. The scalability among the number of homologies and the surrounding humans is implicitly tested in the previous experiment, as they are variable among different time steps among every environment. To test scalability over the trajectory length, we repeat the previous experiment with a higher dataset prediction time horizon to get farther navigation goals. The previous horizon was set to 4.8 s, as it is a common choice in most of the works, and we double it to 9.6 s for this experiment. Note that the network has the same fixed weights (with no fine-tuning or retraining) as before, trained with the short time horizon. The accuracy metrics are shown in Table~\ref{tab:8s}. The accuracy of our method is still significantly higher than the other approaches, due to the fact that the network is not dependent on the trajectory guidance, but on the signature of the homology class, so the weights are not overfitting the trajectory length. 


\begin{table}[ht]
    \caption{Accuracy metrics of the predictions of the human homology class, taking data from the dataset with a prediction horizon of 9.6 s.}
    \centering
    \resizebox{\linewidth}{!}{
    \begin{tabular}{|c|ccccc|c|}
        \hline
        \textbf{Algorithm} & \textbf{ETH} & \textbf{Hotel} & \textbf{Univ} & \textbf{Zara1} & \textbf{Zara2} & \textbf{Avg} \\
        \hline
         $\mathcal{J}_{len}$   & 0.889 & 0.828 & 0.688 & 0.650 & 0.759 & 0.763 \\
         $\mathcal{J}_{acc}$  & 0.722 & 0.797 & 0.697 & 0.676 & 0.783 & 0.735 \\
         $\mathcal{J}_{mix}$   & 0.818 & 0.807 & 0.702 & 0.705 & 0.749 & 0.756 \\
         $\mathcal{J}_{\theta, right}$ (ours)   & 0.727 & 0.720 & 0.614 & 0.764 & 0.620 & 0.689 \\
         $\mathcal{J}_{\theta, ref}$ (ours)     & \textbf{1.000} & \textbf{0.954} & \textbf{0.780} & \textbf{0.789} & \textbf{0.880} & \textbf{0.881} \\
        \hline
    \end{tabular}
    }
    \label{tab:8s}
\end{table}

\subsection{Simulation experiments}\label{sec:simu-eval}

\begin{figure}
    \centering
    \includegraphics[trim={16.5cm 4.4cm 16.5cm 4.5cm},clip,width=0.45\textwidth]{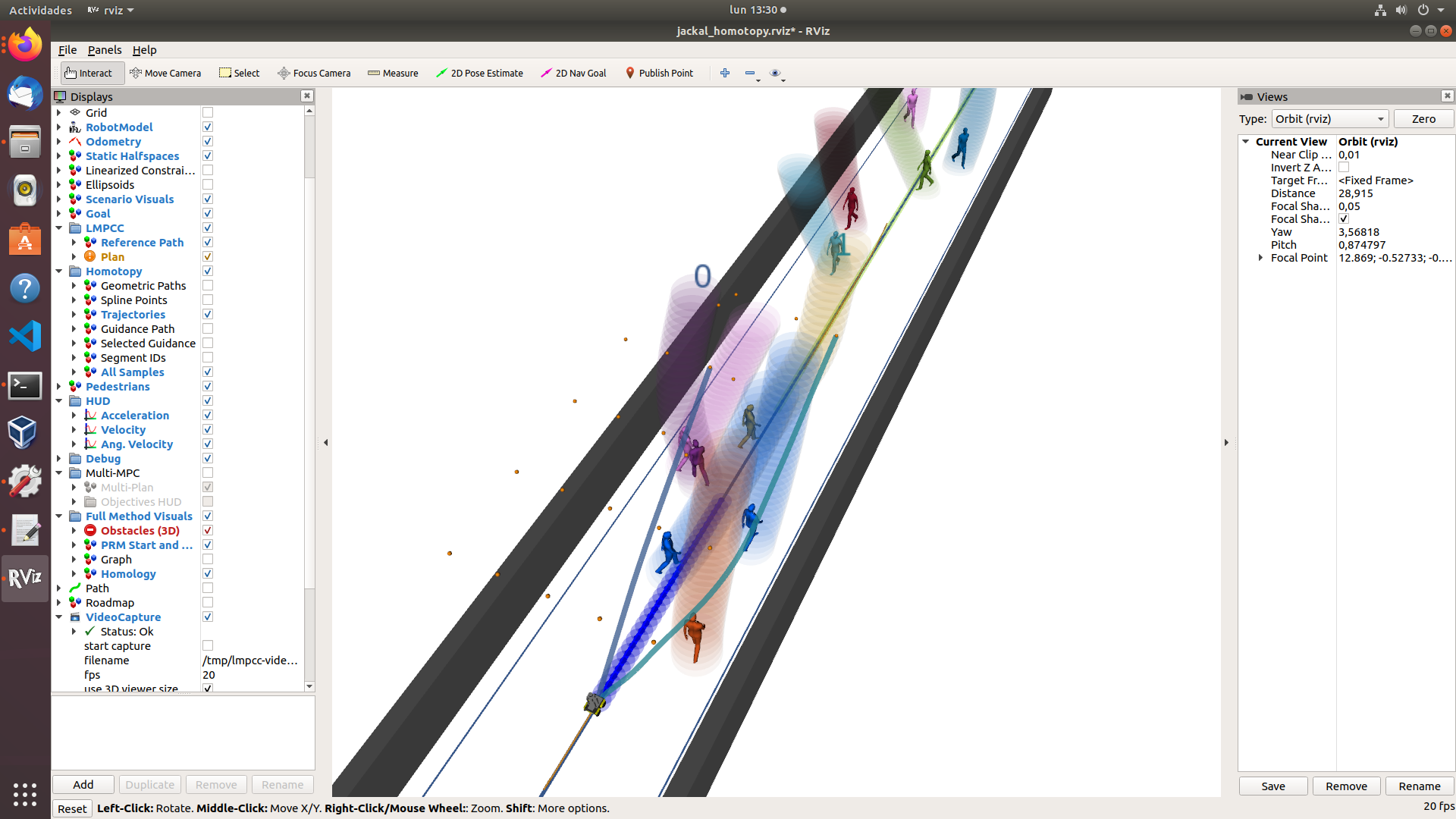}
    \caption{Simulated corridor environment where different planners are tested. The predicted trajectories of the pedestrians are represented with semi transparent shadows of different colors, the plan with an intense blue and two guidance trajectories with two lines that start in the robot.}
    \label{fig:simulator}
\end{figure}

We conducted experiments in a battery of 50 simulation scenarios and gathered behavior metrics, similar to the ones proposed in \citet{biswas2022socnavbench} or \citet{francis2023principles}. The simulated environment (Figure~\ref{fig:simulator}) is a corridor with 12 pedestrians that pass through the corridor and avoid other pedestrians and the robot using Social Force~\citep{helbing1995social}. The simulated robot has differential-drive restrictions, and its task is to navigate through the corridor for 25 m. We gather metrics for a differential drive version of Social Force, a social DRL-based planner \citep{martinez2023improving}, an MPC-based planner (LMPCC) \citep{brito2019model}, and SHINE. As the metrics have different magnitudes, for a better understanding and comparison, we express the metrics related to Social Force, selected as reference. The results are represented in Table~\ref{tab:battery}.

\begin{table}[ht]
    \caption{Social metrics of the different planners. They are expressed in relation to Social Force metrics (except success and collision rates). The arrows indicate whether higher or lower values are preferable.}
    \centering
    \resizebox{\linewidth}{!}{
    \begin{tabular}{|c|cccc|}
        \hline
        \multirow{2}{*}{\textbf{Metric}} & \textbf{Social} & \multirow{2}{*}{\textbf{DRL}} & \multirow{2}{*}{\textbf{LMPCC}} & \multirow{2}{*}{\textbf{SHINE}} \\
        & \textbf{Force} & &  &\\
        \hline
         Success rate $\uparrow$  & 0.74 & 0.90 & \textbf{0.98} & \textbf{0.98}  \\
         Collision rate $\downarrow$  & 0.26 & 0.10 & \textbf{0.02} & \textbf{0.02}  \\
        \hline
         Avg. path length $\downarrow$  & \textbf{1.000} & 1.014 & 1.010 & 1.009  \\
         Avg. time to goal $\downarrow$  & 1.000 & \textbf{0.997} & 1.052  & 1.003  \\
         Avg. speed $\uparrow$  & 1.000 & 1.005 & 0.968 & \textbf{1.010}  \\
         Min. dist. obs. $\uparrow$  & 1.000 & \textbf{1.043} & 0.996 & 1.011  \\
         Avg. dist. obs. $\uparrow$  & 1.000 & 1.020  & \textbf{1.030} & 1.022  \\
         Min. time to coll. $\uparrow$    & 1.000 & \textbf{1.091} & 0.827 & 1.069 \\
         \hline
         Path irregularity $\downarrow$  & 1.000 & 2.093 & \textbf{0.793} & 0.814  \\
         Avg. $\omega$ $\downarrow$  & 1.000 & 2.662 & 0.389 & \textbf{0.311}  \\
         \hline
         Avg. acceleration $\downarrow$  & 1.000 & 1.537 & 1.116 & \textbf{0.805}  \\
         Avg. jerk $\downarrow$  & 1.000 & 1.673 & 1.080 & \textbf{0.760}  \\
        \hline

    \end{tabular}}
    \label{tab:battery}
\end{table}

In these scenarios, both LMPCC-based planners (LMPCC and SHINE) are safer \hh{with higher success rates (number of times the robot completes the task without collisions or getting stuck per scenario) and lower collision rates (number of times the robot collides per scenario). } Social Force's goal is navigating in a way that respects social norms with physics-inspired principles of attractive and repulsive forces, and its simplicity compared to the other planners could contribute to a higher collision rate. The DRL planner was trained in CrowdNav simulator, which uses random positions for humans. Its success rate could increase after training it in corridor scenarios, but it would mean that its effectiveness does not translate to different settings. The metrics of average path length, time to reach the goal, speed, distance to obstacles and time to collision are similar and comparable for all the planners; as their meaning is more related to the local constraints, and we set their parameters to behave similarly.

There is a clear difference in path irregularity (angle between the robot heading and the vector pointing to the goal) and the angular velocity ($\omega$) between the MPC-based planners and the purely reactive ones. The reason why this happens is that these methods plan the whole path in advance, leading to smoother trajectories, which are preferable in social navigation. In addition, there is a difference in average acceleration and jerk (time derivative of acceleration) between our planner and the rest, as it is able to escape from local minima and replan when the environment changes, instead of having to suddenly accelerate or brake to avoid collisions.

\begin{figure}
    \centering
    \begin{tabular}{@{}c@{}}
      \includegraphics[width=0.7\linewidth]{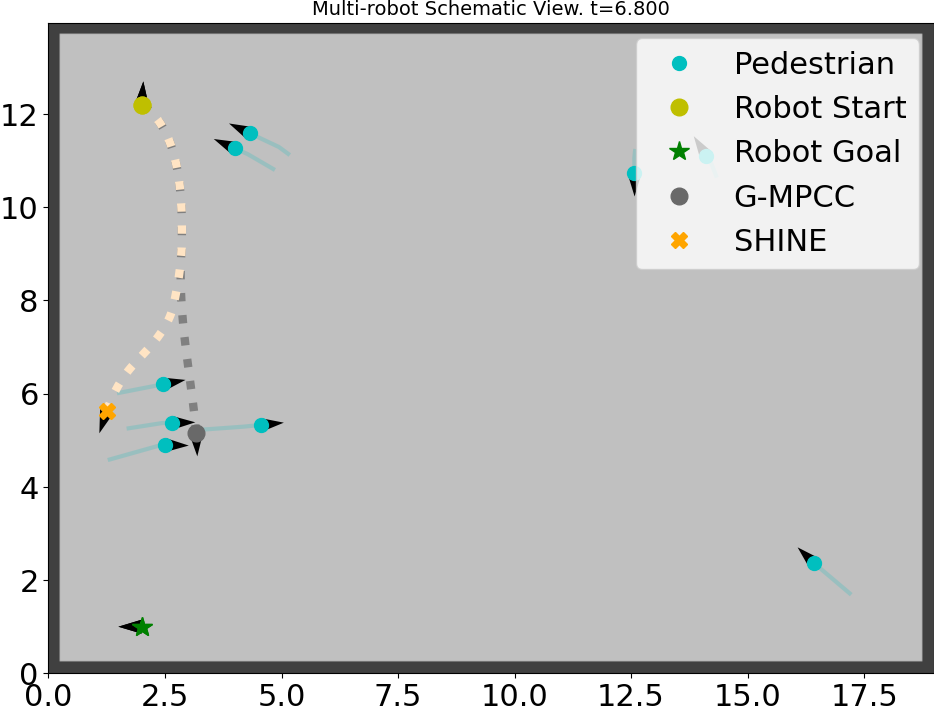}   \\
      \footnotesize{(a)} \\
      \includegraphics[width=0.7\linewidth]{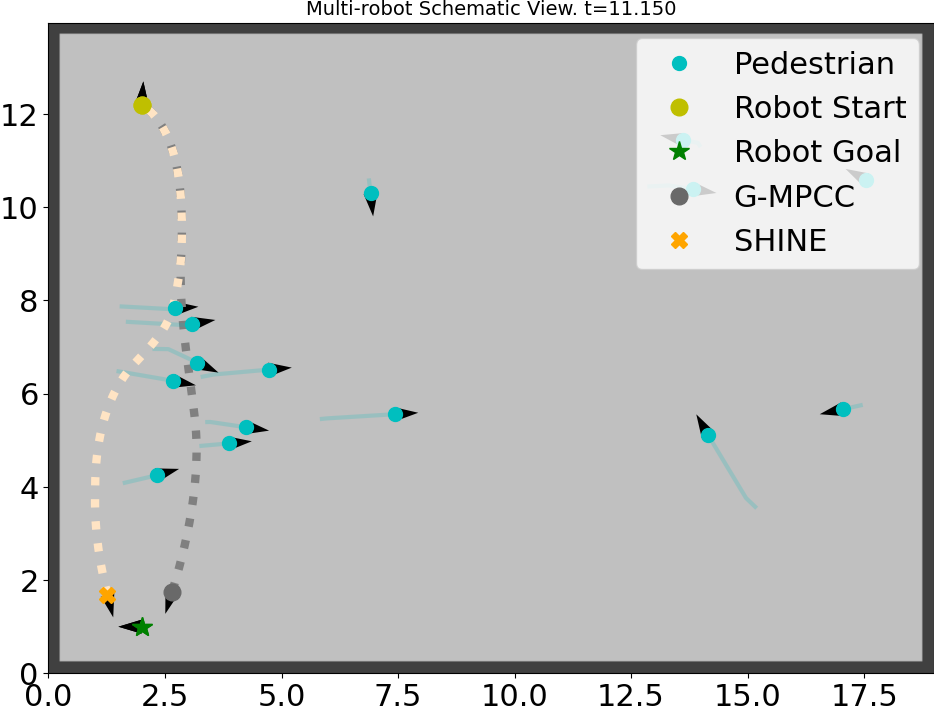}\\
       \footnotesize{(b) }
    \end{tabular}
    \caption{A scenario (named \textit{univ\_trapTL}) of SocNavBench showing SHINE and G-MPCC with $\mathcal{J}_{acc}$ cost trajectories.}
    \label{fig:socnavbench-example}
\end{figure}

\hh{Finally, we conducted experiments to assess the benefit of learning the guidance trajectories from humans rather than designing a hand-crafted cost for it. To do so, we used SocNavBench~\citep{biswas2022socnavbench}, a simulator that generates humans using prerecorded trajectories, to compare SHINE with Guidance-MPCC using $\mathcal{J}_{len}$, $\mathcal{J}_{acc}$ and $\mathcal{J}_{mix}$. In this way, we intend to have more realistic simulations that resemble the real world and to use an existing benchmark. The metrics gathered for 5 runs of the 33 scenarios proposed as a benchmark in SocNavBench are shown in Table~\ref{tab:socnavbench}. We do not include comparisons with other planners of the simulator, as they do not consider robot dynamics. In this simulator, humans are spawned at the time when the recorded data of each specific human starts. Therefore, humans are sometimes spawned right into the robot position, resulting in an unavoidable collision and not reliable metrics. In addition, some scenarios may not have trajectories in different homology classes as either the robot faces no obstacles in its way to the goal or the scenario is too crowded. Therefore, differences in Table~\ref{tab:socnavbench} among the methods are small. Thus, we selected the benchmark scenarios where SHINE and $\mathcal{J}_{acc}$, which has the best performance in Table~\ref{tab:socnavbench} among the hand-crafted costs, navigated within different homology classes without collisions. The resulting metrics in those eleven scenarios are shown in Table~\ref{tab:socnavbench-homo}. The results prove a better performance of SHINE with respect any other alternative used as baseline when it chooses a different homology class. One example is shown in Fig.~\ref{fig:socnavbench-example}, where both G-MPCC and SHINE reach the goal succesfully at similar times, but SHINE in a way that is less intrusive regarding pedestrians. The metrics show that SHINE exhibits safety with high success rates, efficiency with low task duration and path length ratio metrics, as well as higher minimum distance to pedestrians and minimum time to collision metrics per episode. It is specially remarkable how it reaches the goal in considerably less time while keeping the highest time to collision to pedestrians. }

\begin{table}[ht]
    \caption{\hh{Metrics gathered in SocNavBench.}}
    \centering
    \resizebox{\linewidth}{!}{
    \begin{tabular}{|c|cccc|}
        \hline
        \hh{\textbf{Metric}} & \hh{\textbf{$\mathcal{J}_{len}$}} & \hh{\textbf{$\mathcal{J}_{acc}$}} & \hh{\textbf{$\mathcal{J}_{mix}$}} & \hh{\textbf{SHINE}}\\
        \hline
\hh{Success rate $\uparrow$} & \hh{0.850 }& \hh{\textbf{0.912} }& \hh{0.856 }& \hh{\textbf{0.912} }\\
\hh{Duration (s) $\downarrow$} & \hh{12.676 (2.7)}& \hh{12.211 (2.5)}& \hh{12.663 (2.7)}& \hh{\textbf{11.994} (2.6)}\\
\hh{Path lenght ratio $\downarrow$} & \hh{0.976 (0.0)}& \hh{0.964 (0.0)}& \hh{0.976 (0.0)}& \hh{\textbf{0.957} (0.0)}\\
\hh{Min. closest ped. dist. (m) $\uparrow$} & \hh{0.880 (0.9)}& \hh{0.942 (0.8)}& \hh{0.879 (0.8)}& \hh{\textbf{0.945} (0.8)}\\
\hh{Min. time to coll. $\uparrow$} & \hh{1.065 (2.1)}& \hh{1.114 (2.0)}& \hh{1.039 (2.1)}& \hh{\textbf{1.141} (2.0)}\\
\hline
    \end{tabular}
    }
    \label{tab:socnavbench}
    \end{table}

\begin{table}[ht]
    \caption{\hh{Metrics gathered in SocNavBench in scenarios where SHINE and Guidance-MPCC with $\mathcal{J}_{acc}$ are in different homology classes.}}
    \centering
    \resizebox{\linewidth}{!}{
    \begin{tabular}{|c|cccc|}
        \hline
        \hh{\textbf{Metric}} & \hh{\textbf{$\mathcal{J}_{len}$}} & \hh{\textbf{$\mathcal{J}_{acc}$}} & \hh{\textbf{$\mathcal{J}_{mix}$}} & \hh{\textbf{SHINE}}\\
        \hline
\hh{Success rate $\uparrow$} & \hh{0.909 }& \hh{\textbf{1.000} }& \hh{0.909 }& \hh{\textbf{1.000} }\\
\hh{Duration (s) $\downarrow$} & \hh{13.735 (3.4)}& \hh{13.223 (2.1)}& \hh{13.740 (2.7)}& \hh{\textbf{12.527} (2.6)}\\
\hh{Path lenght ratio $\downarrow$} & \hh{0.993 (0.0)}& \hh{0.992 (0.0)}& \hh{1.016 (0.1)}& \hh{\textbf{0.972} (0.0)}\\
\hh{Min. closest ped. dist. (m) $\uparrow$} & \hh{0.523 (0.0)}& \hh{0.522 (0.0)}& \hh{0.517 (0.0)}& \hh{\textbf{0.523} (0.0)}\\
\hh{Min. time to coll. $\uparrow$} & \hh{0.401 (0.2)}& \hh{0.368 (0.2)}& \hh{0.360 (0.2)}& \hh{\textbf{0.444} (0.3)}\\
\hline
    \end{tabular}
    }
    \label{tab:socnavbench-homo}
    \end{table}

\subsection{Real world experiments}\label{sec:real-eva}

We evaluated our approach in the real platform and compared it with other methods of the state of the art. \hh{We conducted two kind of experiments. First, a group of experiments to assess the limitations of other methods and the behavior of SHINE in those situations. We designed simple scenarios which could make planners fail due to their design.} Particularly, we observed problems of Social Force in what we called push out scenarios, Social DRL in scenarios different from what experienced in training and LMPCC in zig-zag scenarios. \hh{The second type of experiments were conducted to} test the navigation capabilities in crowded environments. In every experiment, the robot is placed in a corner of the room and has to navigate to the other corner. A visual summary of the real-world experiments is recorded in the supplementary video\hh{, which may be accessed in the project repository~\citep{martinez2023github}}. 

\subsubsection{Push out scenario}

\begin{figure}
    \centering
    \begin{tabular}{@{}cc@{}}
     \includegraphics[width=0.22\textwidth]{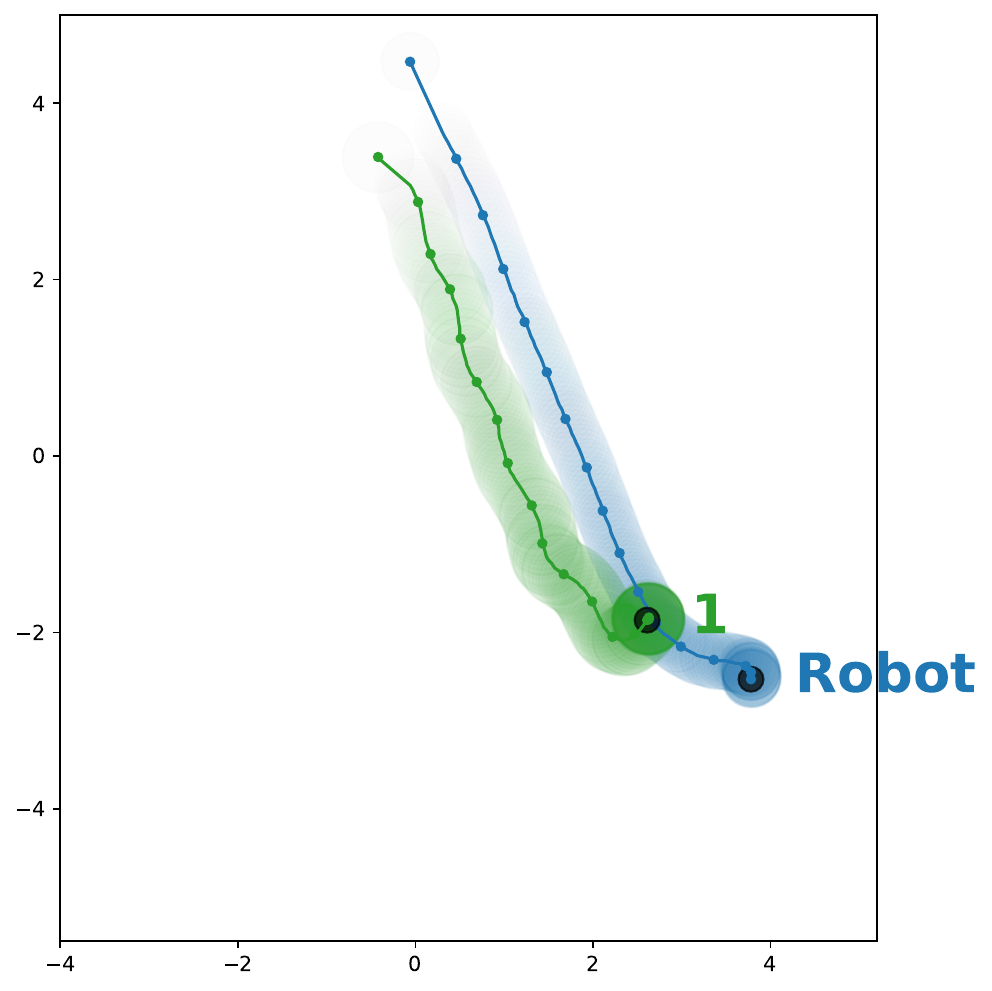} & \includegraphics[width=0.22\textwidth]{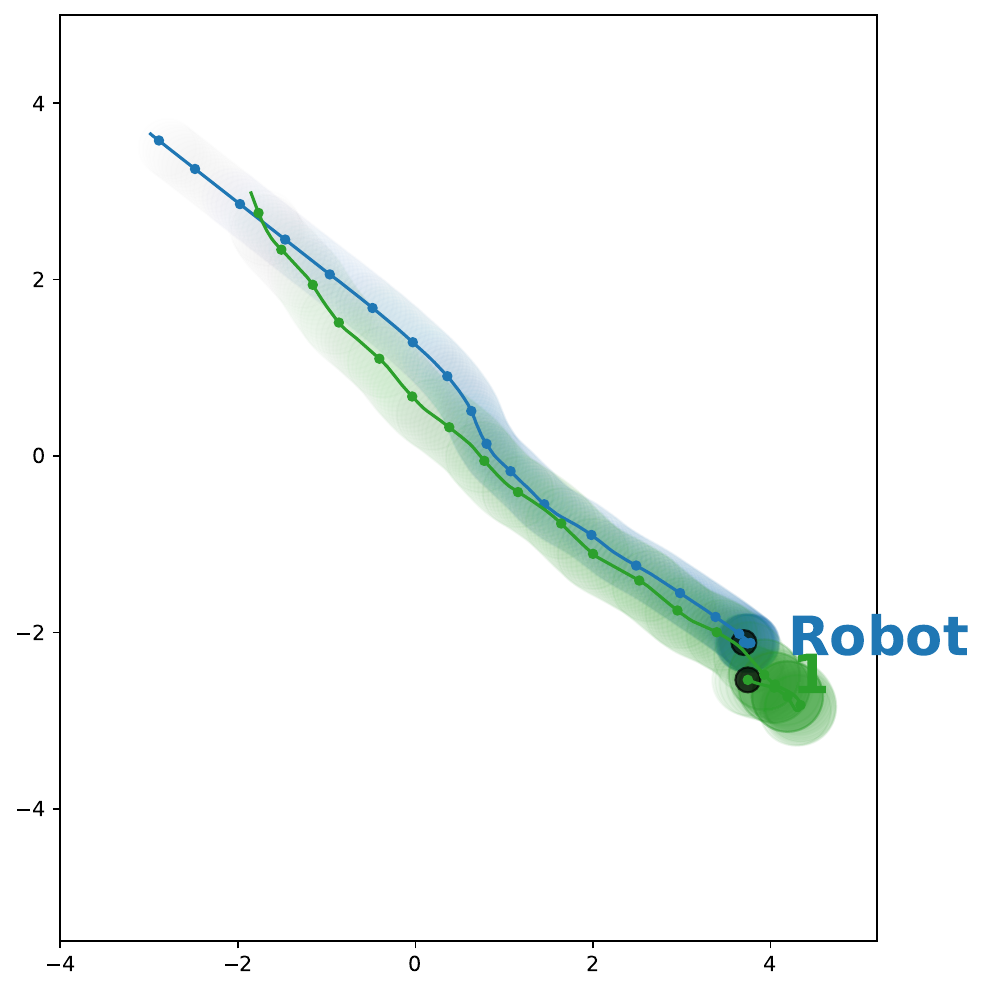} \\
       \footnotesize{(a) Social Force}  & \footnotesize{(b) SHINE}
    \end{tabular}
    \caption{Graphs of the result of a push out scenario with Social Force (a) and SHINE (b) planners. The robot is represented in blue and humans in green. The starting positions are represented with a black circle and the trajectory in time with a shadow with increasing transparency.}
    \label{fig:pushout}
\end{figure}

We tried an experiment where a human walks in the same direction the robot is trying to avoid it, closing its way. Social Force~\citep{helbing1995social} exhibits limitations in performance in scenarios where the attractive and repulsive forces are opposing, as in Figure~\ref{fig:pushout} (a). Thus, the resulting behavior of this planner is that the robot keeps trying to avoid the human in that way and, therefore, the human pushes the robot out of the original way. In these situations, SHINE makes the robot brake or avoid the human through the other side, since it plans a long-term trajectory (Figure~\ref{fig:pushout} (b)).

\subsubsection{Static scenario} 
\begin{figure}
    \centering
    \begin{tabular}{@{}cc@{}}
     \includegraphics[width=0.22\textwidth]{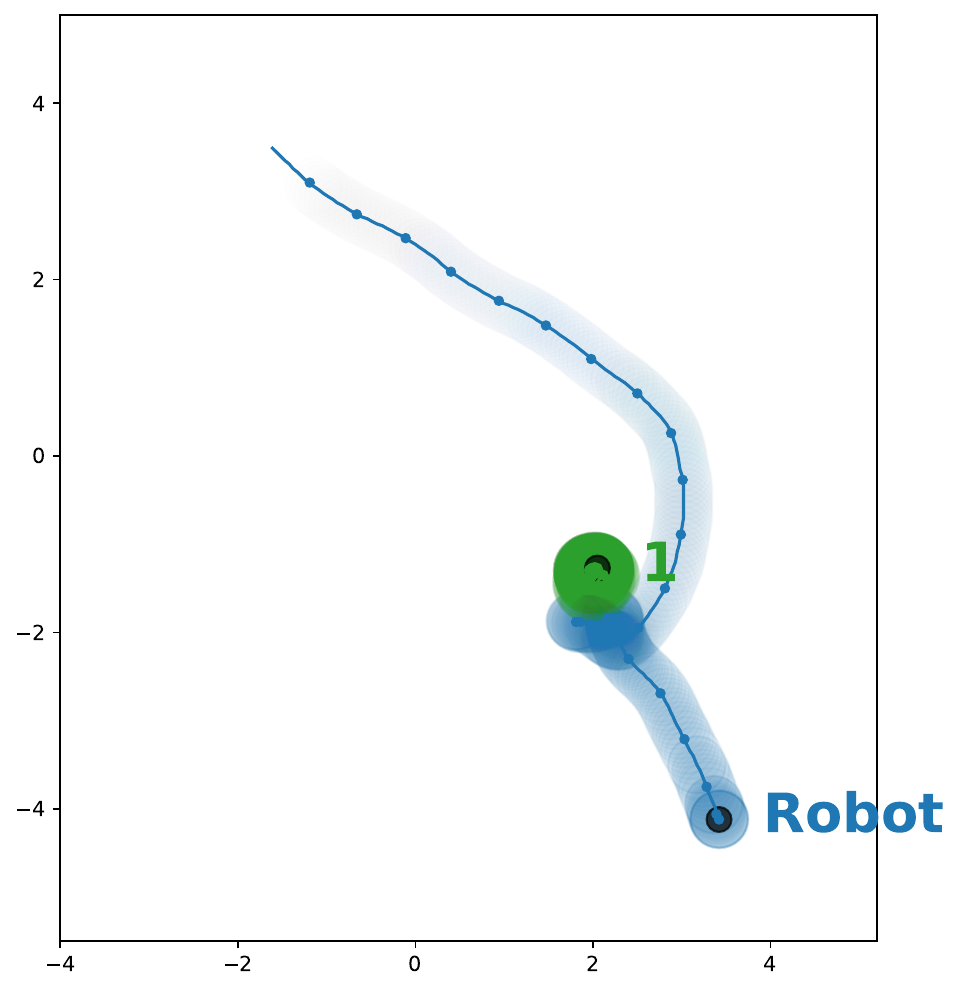} & \includegraphics[width=0.22\textwidth]{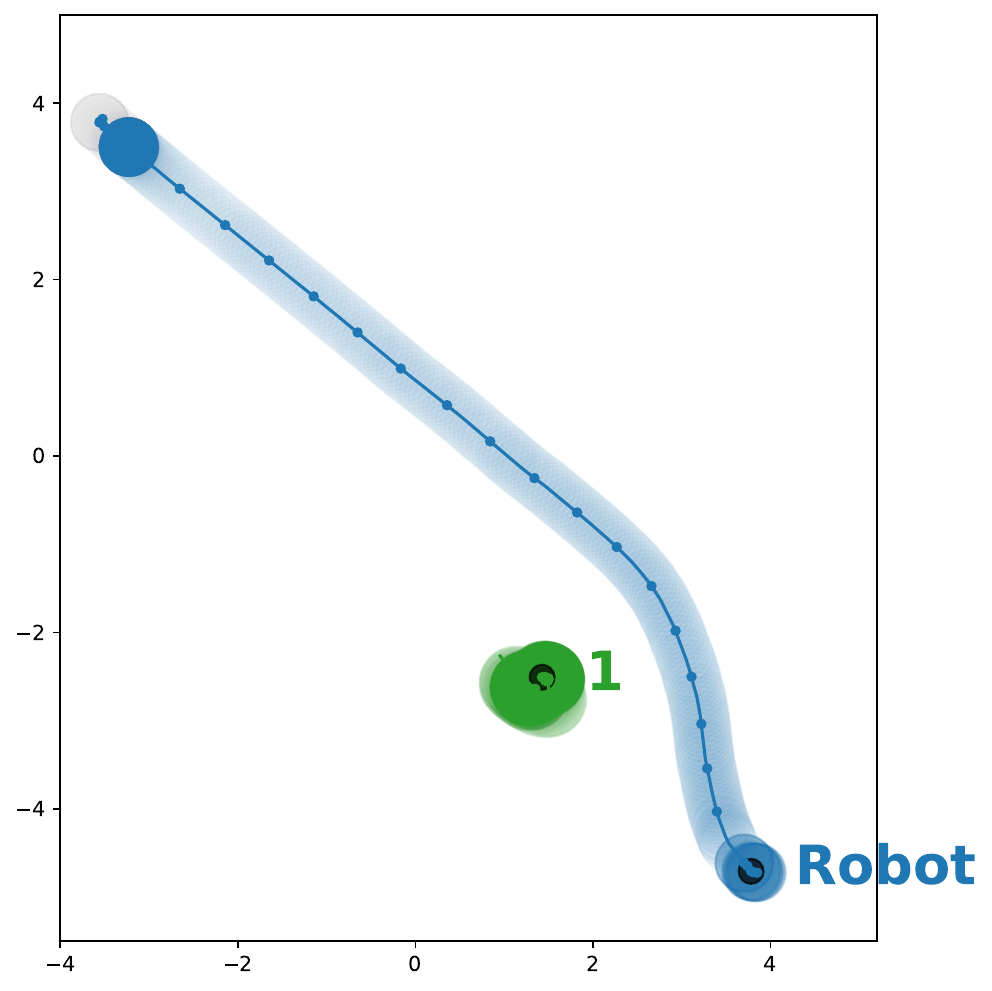} \\
       \footnotesize{(a) DRL}  & \footnotesize{(b) SHINE}
    \end{tabular}
    \caption{Graphs of the result of a static scenario with Social DRL (a) and SHINE (b) planners. The figure follows the same representation as Figure~\ref{fig:pushout}.}
    \label{fig:static}
\end{figure}

End-to-end DRL methods reactively send the robot the velocity commands they estimate to provide the highest reward. Their limitations are that they do not provide guarantees for collision avoidance, they are reactive, and their performance is restricted to what they have learned. We included a DRL algorithm \citep{martinez2023improving} in our platform and detected that it had room for improvement in scenarios where humans remained static. In these situations, it reacted late (possibly expecting motion from the human side). This is probably due to the fact that in the simulator it was trained, mostly dynamic obstacles were considered. Even though the performance could be probably enhanced with some refinements in the training process, it is an example of the challenges DRL faces nowadays. SHINE has collision restrictions and plans in advance, so it may adapt to any environments. The difference in performance is shown in Figure~\ref{fig:static} (a) and Figure~\ref{fig:static} (b).

\subsubsection{Zig-zag scenario} 
\begin{figure}
    \centering
    \begin{tabular}{@{}cc@{}}
     \includegraphics[width=0.22\textwidth]{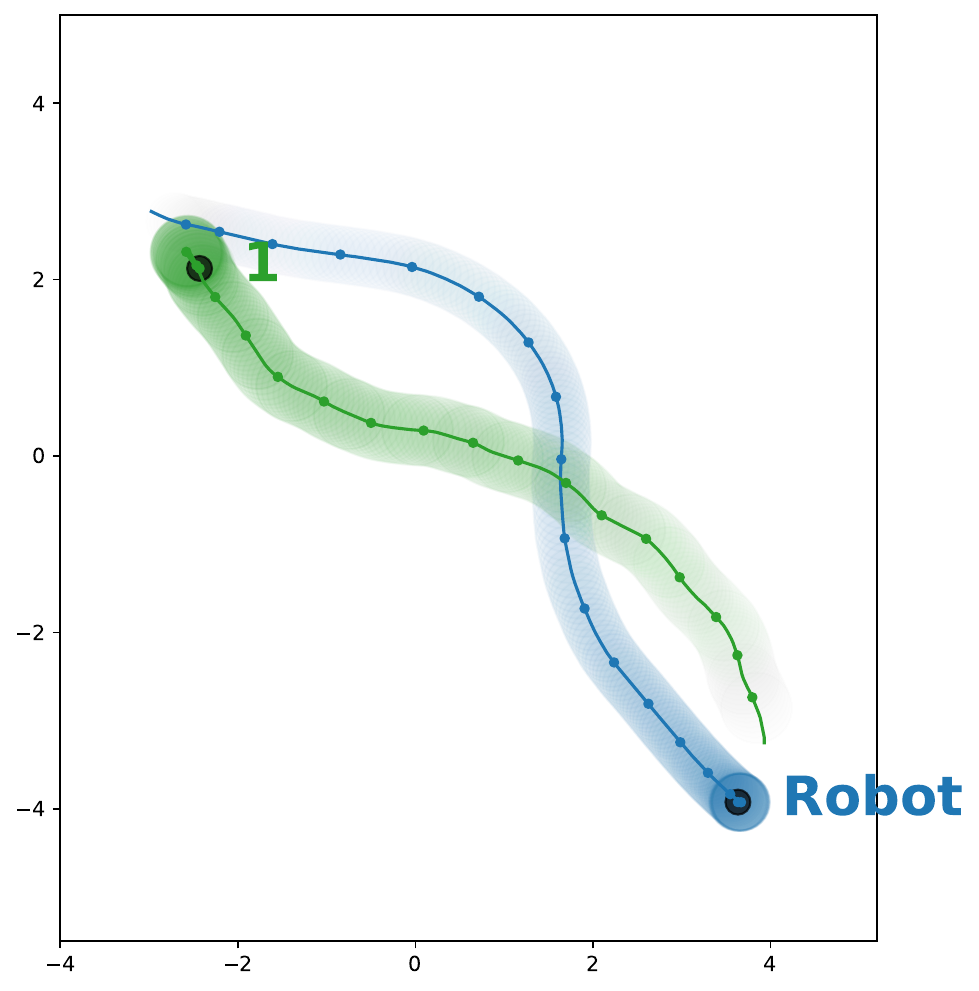} & \includegraphics[width=0.22\textwidth]{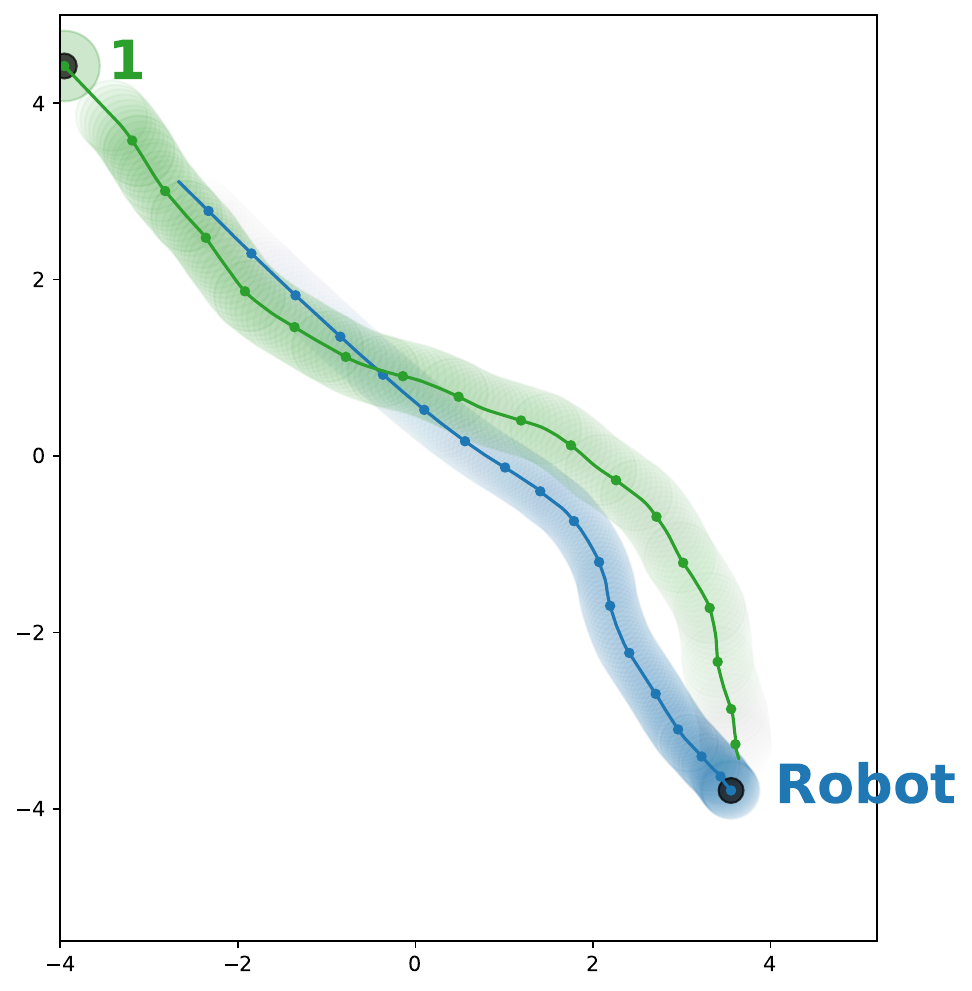} \\
       \footnotesize{(a) LMPCC}  & \footnotesize{(b) SHINE}
    \end{tabular}
    \caption{Graphs of the result of a zig-zag scenario with LMPCC (a) and SHINE (b) planners. The figure follows the same representation as Figure~\ref{fig:pushout}.}
    \label{fig:boomerang}
\end{figure}

An MPC-based method gets trapped in local optima if the predictions do not meet the real behavior of the obstacles. We tested LMPCC \citep{brito2019model} in a scenario where a pedestrian chooses one side to avoid the robot and then, unexpectedly, tries to avoid it through the other side, following a "zig-zag" motion. The result, as seen in Figure~\ref{fig:boomerang} (a), is that the robot tries to keep the same homology class, accelerating and curving its trajectory to avoid the human through the same side (it passes before the pedestrian). Our approach (Figure~\ref{fig:boomerang} (b)) replans in a different homology class to avoid the pedestrian to avoid the pedestrian through the other side (it passes after the pedestrian), resulting in a smoother and less curvy trajectory.

\subsubsection{Crowded scenarios} 

\hh{This experiment differed from the others in that we were not addressing a specific limitation of a method but rather assessing overall performance. We conducted the experiment in scenarios involving several moving pedestrians. Five participants were randomly selected and instructed to wander around the room freely. They were informed that a robot would also be moving in the room and that they should assume it would behave like another person.}

\hh{To begin, participants were asked to walk around the room for one minute without the robot, allowing them to get accustomed to the environment. Following this, they repeated the exercise while sharing the space with the moving robot for five minutes. Afterward, each participant individually and separately evaluated the robot's motion qualitatively for one minute to avoid mutual influence.}

\hh{This sequence was repeated four times, each time with a different motion planner controlling the robot. The planners tested in the experiment were Social Force, DRL, Guidance-MPCC, and SHINE, in that order. We recorded the positions of both the robot and the pedestrians, participants' evaluations of each planner, and a video of the experiment.}

\hh{In the evaluations, all the participants} assigned higher grades to SHINE than to Guidance-MPCC. Social Force and DRL had a very reactive behavior that was not desirable for navigating around humans, even leading to some collisions.

\begin{figure}
    \centering
    \begin{tabular}{@{}cc@{}}
     \includegraphics[width=0.22\textwidth]{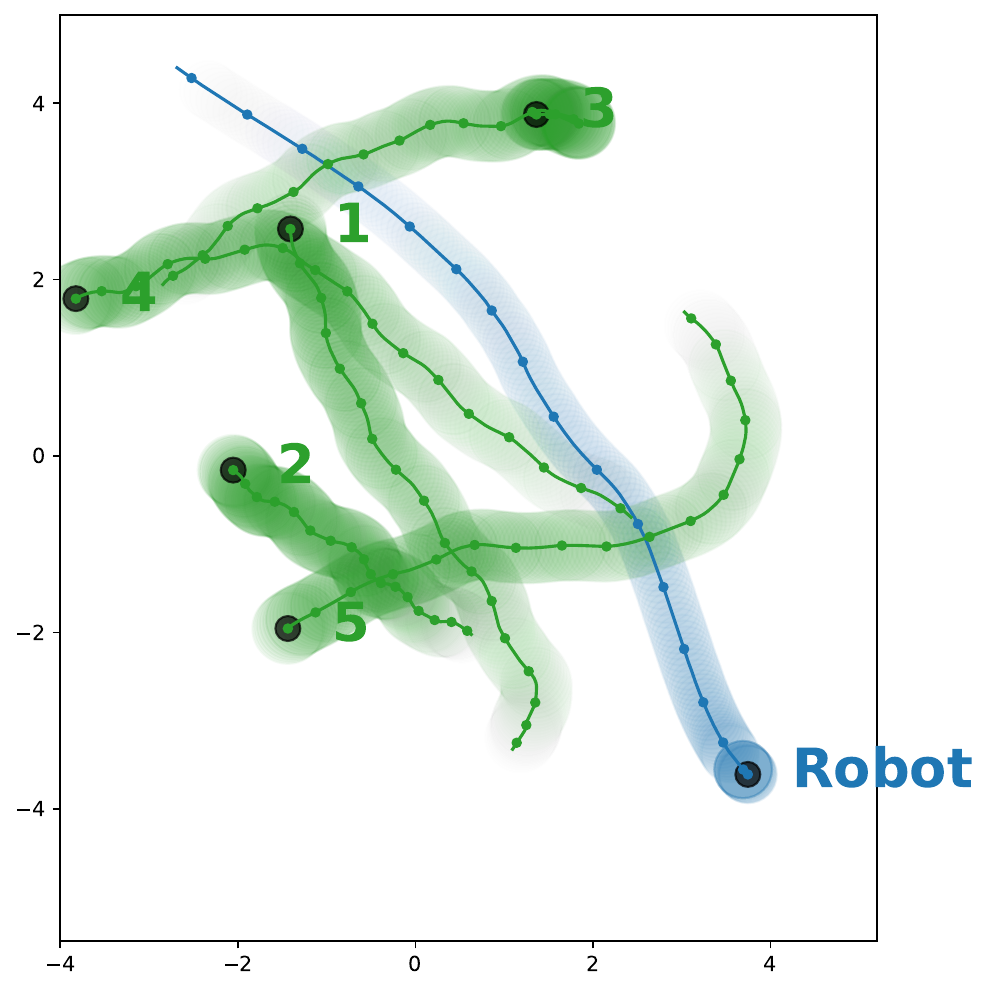} & \includegraphics[width=0.22\textwidth]{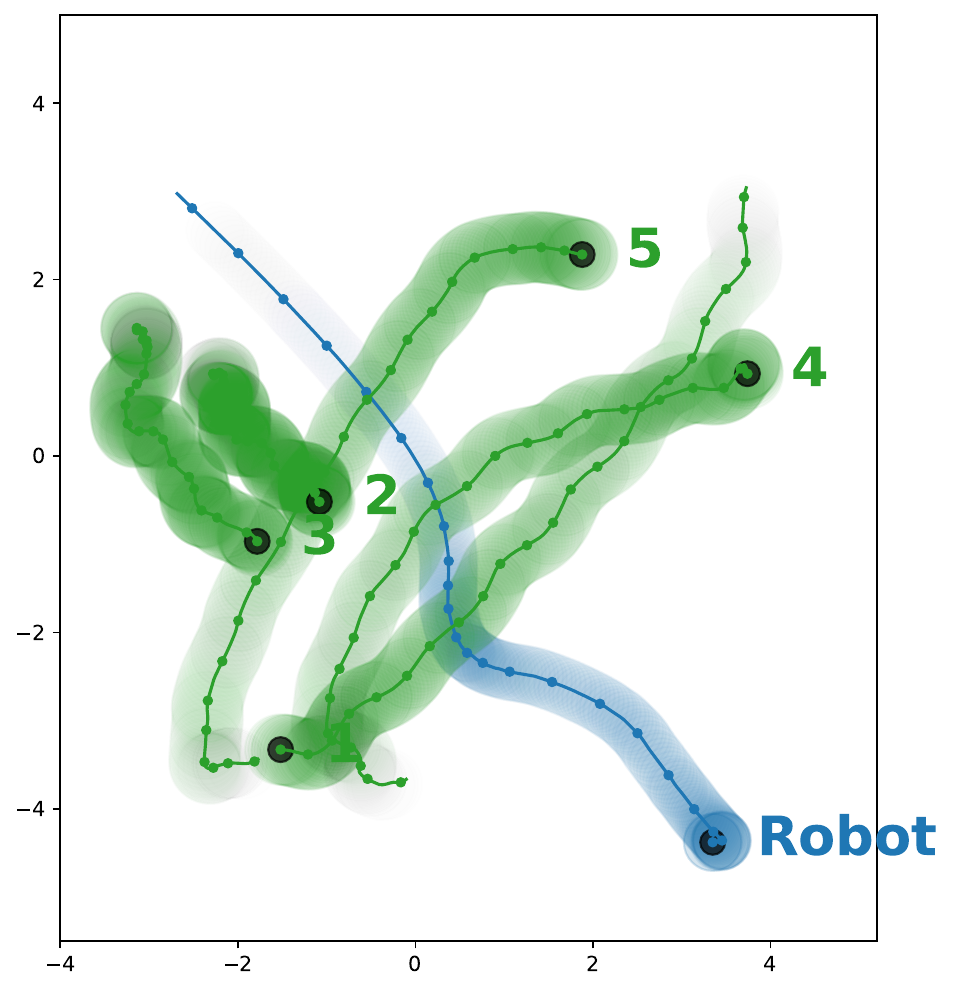} \\
       \footnotesize{(a) Guidance-MPCC}  & \footnotesize{(b) SHINE}
    \end{tabular}
    \caption{Graphs of the result of a crowded scenario with Guidance-MPCC (a) and SHINE (b) planners. The figure follows the same representation as Figure~\ref{fig:pushout}. Guidance-MPCC decides to pass before the first human it encounters (a), making him adapt its trajectory, while SHINE passes after him (b), so the resulting trajectory of the human is smooth.}
    \label{fig:5-peds-real}
\end{figure}

We analyzed the behavior of SHINE in the experiments and compared it with the behavior of Guidance-MPCC in terms of the homology class chosen in Table~\ref{tab:pass-beh}. In every situation where the robot could choose among different homology classes, we annotated how many times it chose to avoid the corresponding obstacle through the right or left. Additionally, in situations that were not head-ons, we measured the number of times the robot chose to pass before or after the corresponding human. The table shows that there is a higher tendency of avoiding on the right side in both methods. This probably happens because the pedestrians participating in the experiment have a cultural tendency of choosing the right side. Nevertheless, when comparing the times the robot chose to pass before or after the obstacle, there is a clear difference. Guidance-MPCC prefers passing before other pedestrians, as it chooses the guidance trajectories where it may reach the goal faster while keeping high and constant velocities. Our network, however, has a clear preference of passing after the other humans, due to the fact that it is socially desirable to not interrupt other pedestrians' trajectories if they have already started them unless there is enough space and they are not walking fast. For example, Figure~\ref{fig:5-peds-real} (a) shows a scenario where the robot navigating with Guidance-MPCC explicitly maneuvers to accelerate and pass before the first human it encounters. Human 5, initially on a direct path, was forced to adopt a curvy trajectory, even though his intended trajectory was clear in advance. Figure~\ref{fig:5-peds-real} (b) shows how SHINE maneuvers to pass after the first human it faces, so the human does not have to modify its trajectory and it looks smooth, resulting in higher perceived comfort.

\begin{table}[ht]
    \caption{Passing behavior of Guidance-MPCC and SHINE in the real-world experiment.}
    \centering
    \begin{tabular}{|c|cccc|}
        \hline
        \textbf{Method} & \textbf{Left} & \textbf{Right} & \textbf{Before} & \textbf{After} \\
        \hline
         G-MPCC & 0.437 & 0.563 & 0.596 & 0.404 \\
         SHINE & 0.463 & 0.538 & \textbf{0.329} & \textbf{0.671} \\
        \hline
    \end{tabular}

    \label{tab:pass-beh}
\end{table}

\subsubsection{Scenarios analysis} 
\begin{figure}
    \centering
    \begin{tabular}{@{}cc@{}}
         \includegraphics[trim={13.5cm 0.4cm 13.5cm 4.5cm},clip,width=0.22\textwidth]{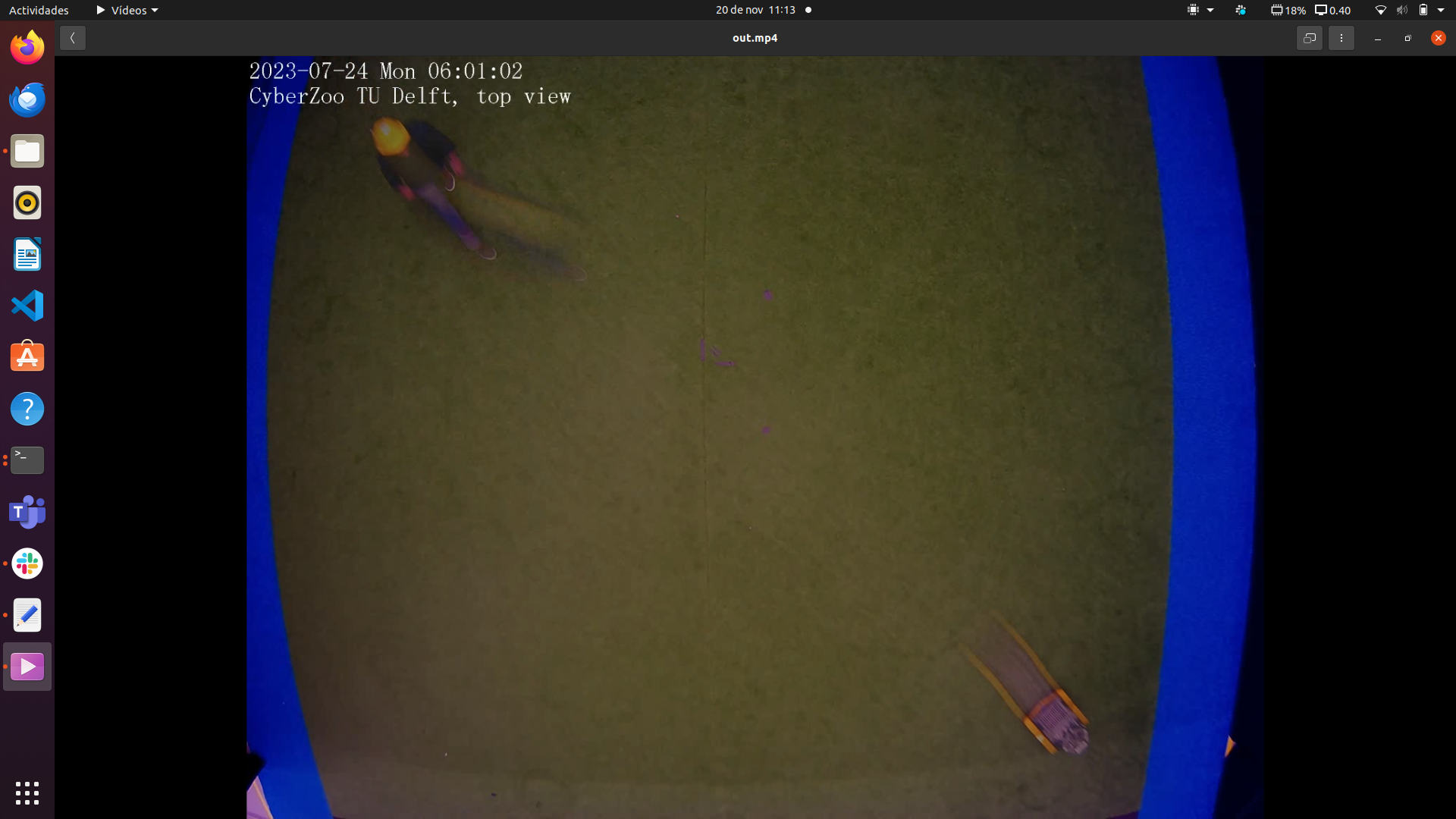}  & 
         \includegraphics[trim={8.43cm 1.09cm 3.94cm 2.53cm},clip,width=0.22\textwidth]{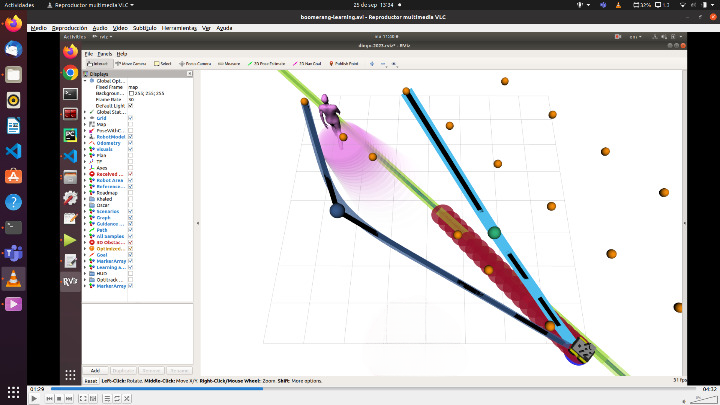}\\
         \footnotesize{(a)} & 
        \footnotesize{(d)} \\
        \includegraphics[trim={13.5cm 0.4cm 13.5cm 4.5cm},clip,width=0.22\textwidth]{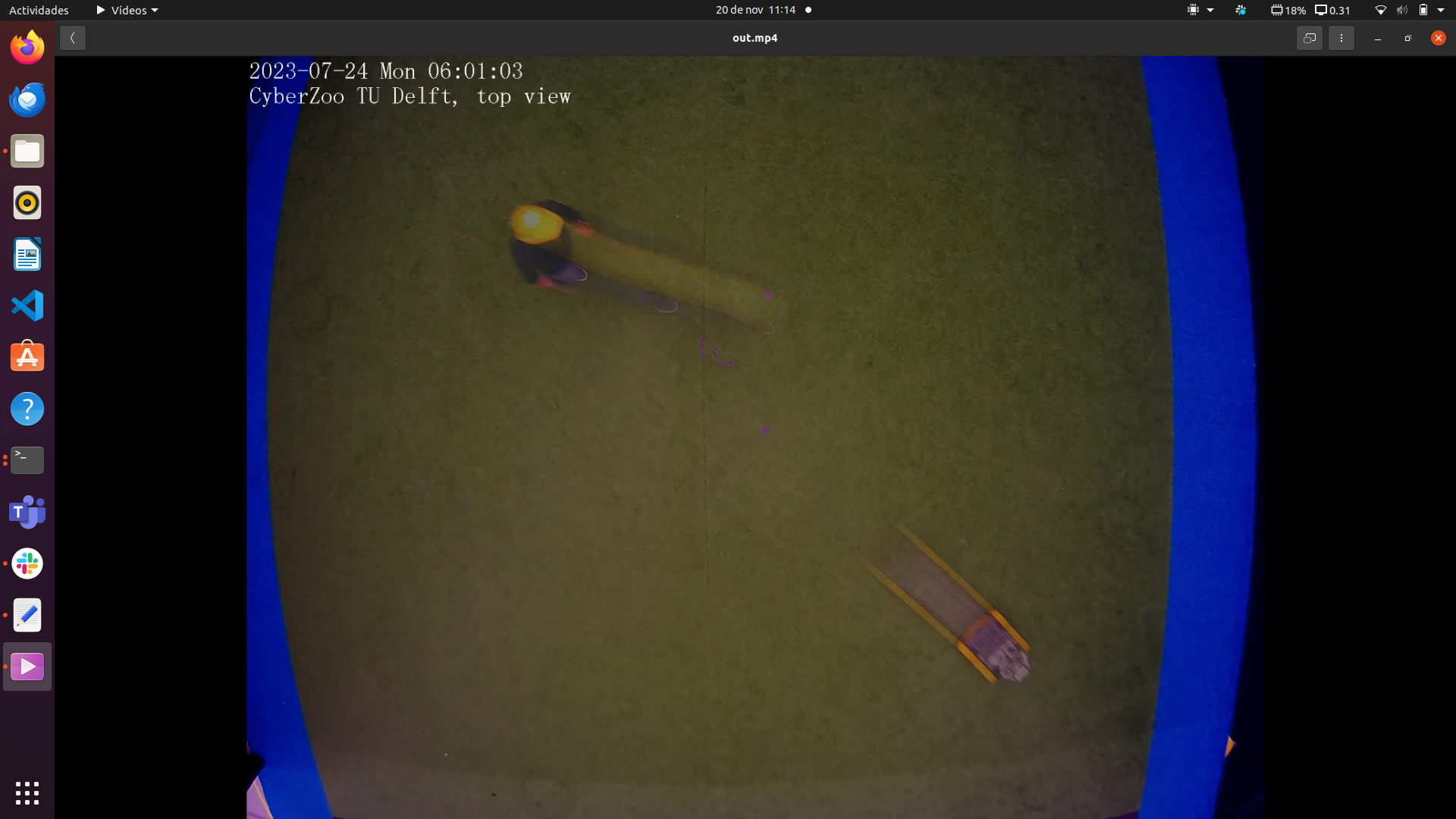}  & 
         \includegraphics[trim={8.43cm 1.09cm 3.94cm 2.53cm},clip,width=0.22\textwidth]{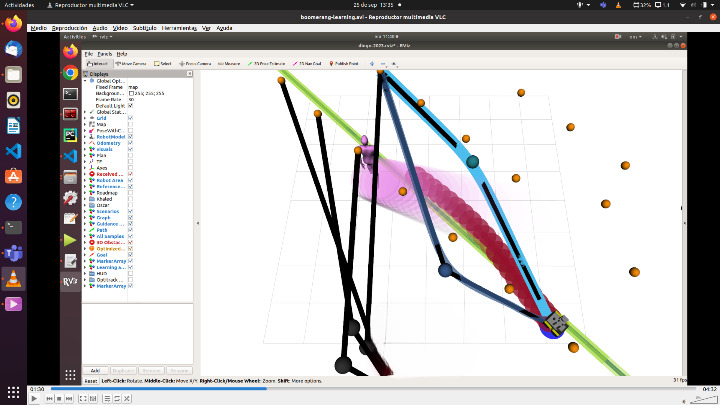}\\
         \footnotesize{(b)} & 
        \footnotesize{(e)} \\
        \includegraphics[trim={13.5cm 0.4cm 13.5cm 4.5cm},clip,width=0.22\textwidth]{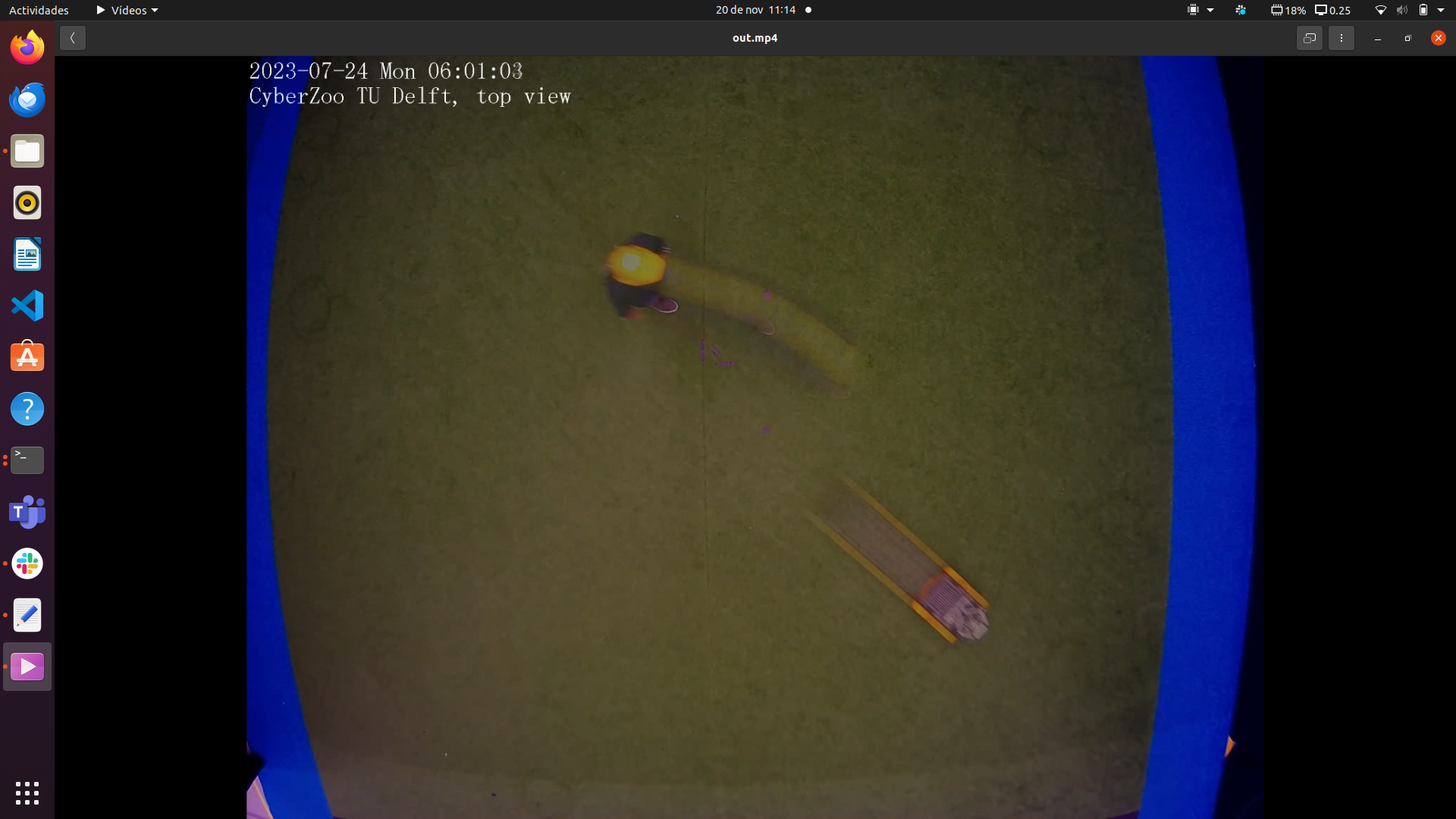}  & 
         \includegraphics[trim={8.43cm 1.09cm 3.94cm 2.53cm},clip,width=0.22\textwidth]{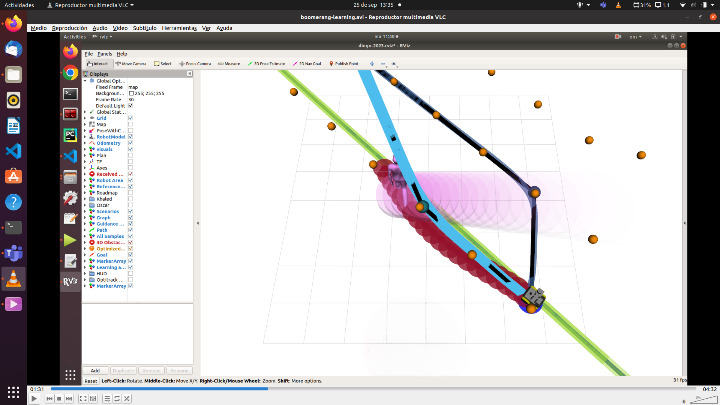}\\
         \footnotesize{(c)} & 
        \footnotesize{(f)} \\
    \end{tabular}
    \caption{Video frames and visualization of a sequence where a robot adapts reactively to human intentions. The light green line that crosses the visualization is the reference path to be followed by the robot. The shadow with increasing transparency that starts in the human is its predicted trajectory. The orange dots are the possible subgoals to follow the reference paths. The colored lines that connect the robot and the subgoals are possible guidance trajectories, while the one highlighted with light blue is the one chosen. The red circles on the ground represent the optimized path computed by LMPCC.}
    \label{fig:change-homo}
\end{figure}

An interesting behavior of our system may be observed in Figure~\ref{fig:change-homo}, where there is a head-on scenario where a person changes the way of avoiding the robot (zig-zag scenario). At first, the human's intention is avoiding the robot on the right side, and the guidance chosen by the robot avoids the human on the right side too (light blue). Then, the human decides to avoid the robot through the left. The robot perceives his intentions and changes the homology class, following a smooth avoidance maneuver.

\begin{figure}
    \centering
    \begin{tabular}{@{}cc@{}}
         \includegraphics[trim={13.5cm 0.4cm 13.5cm 4.5cm},clip,width=0.22\textwidth]{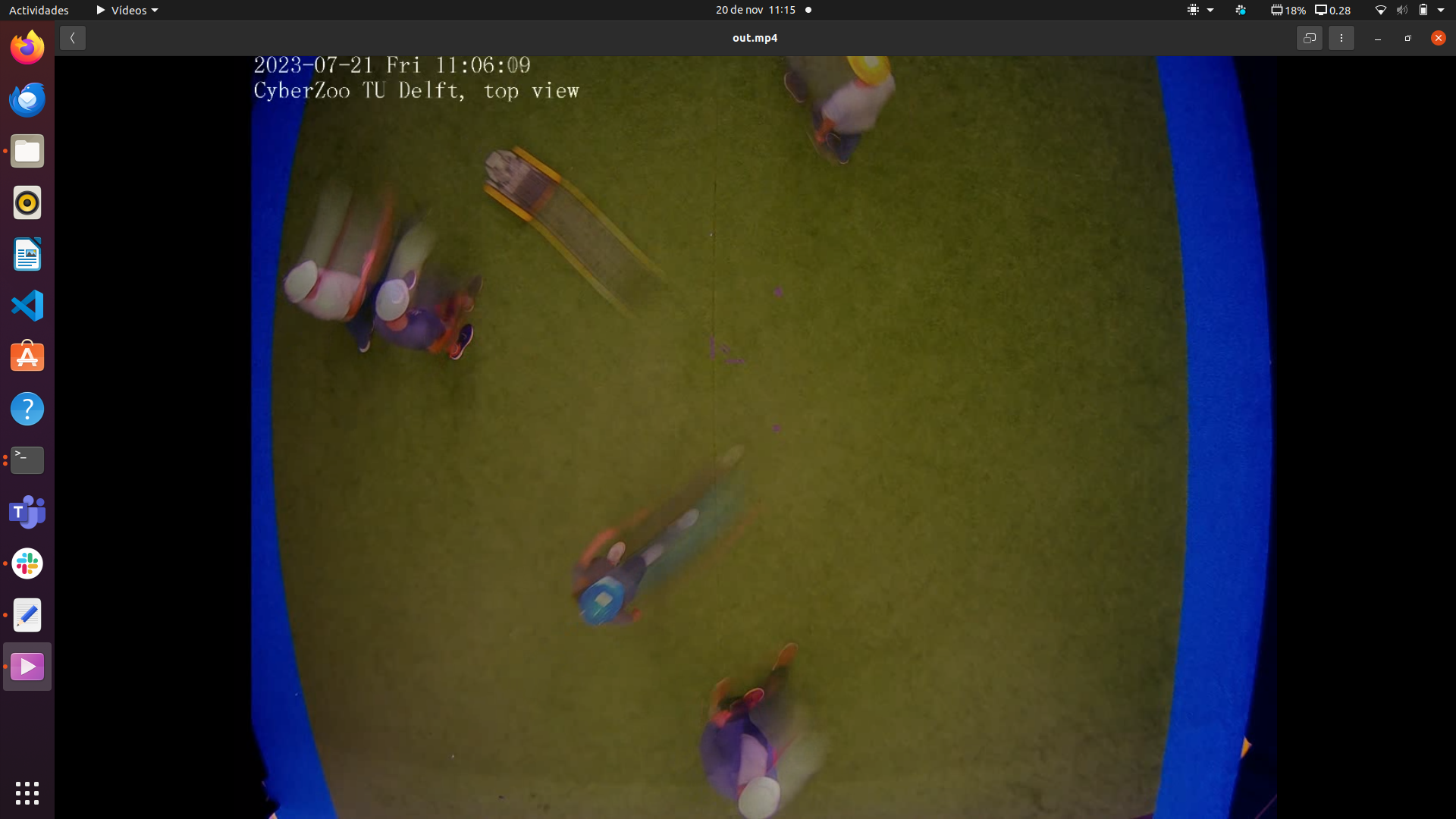}  & 
         \includegraphics[trim={22.5cm 2.9cm 10.5cm 6.75cm},clip,width=0.22\textwidth]{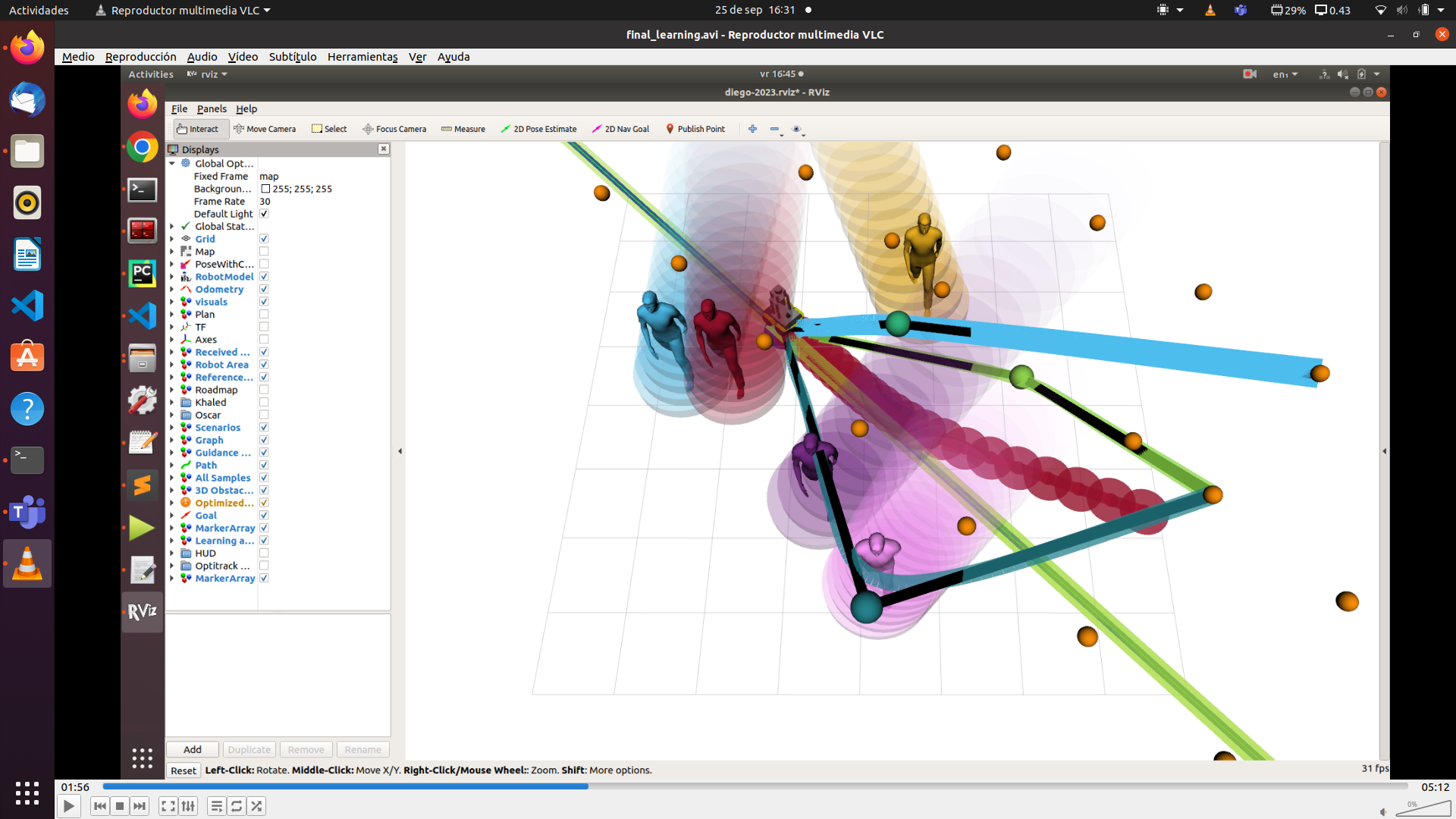}\\
         \footnotesize{(a)} & 
        \footnotesize{(d)} \\
        \includegraphics[trim={13.5cm 0.4cm 13.5cm 4.5cm},clip,width=0.22\textwidth]{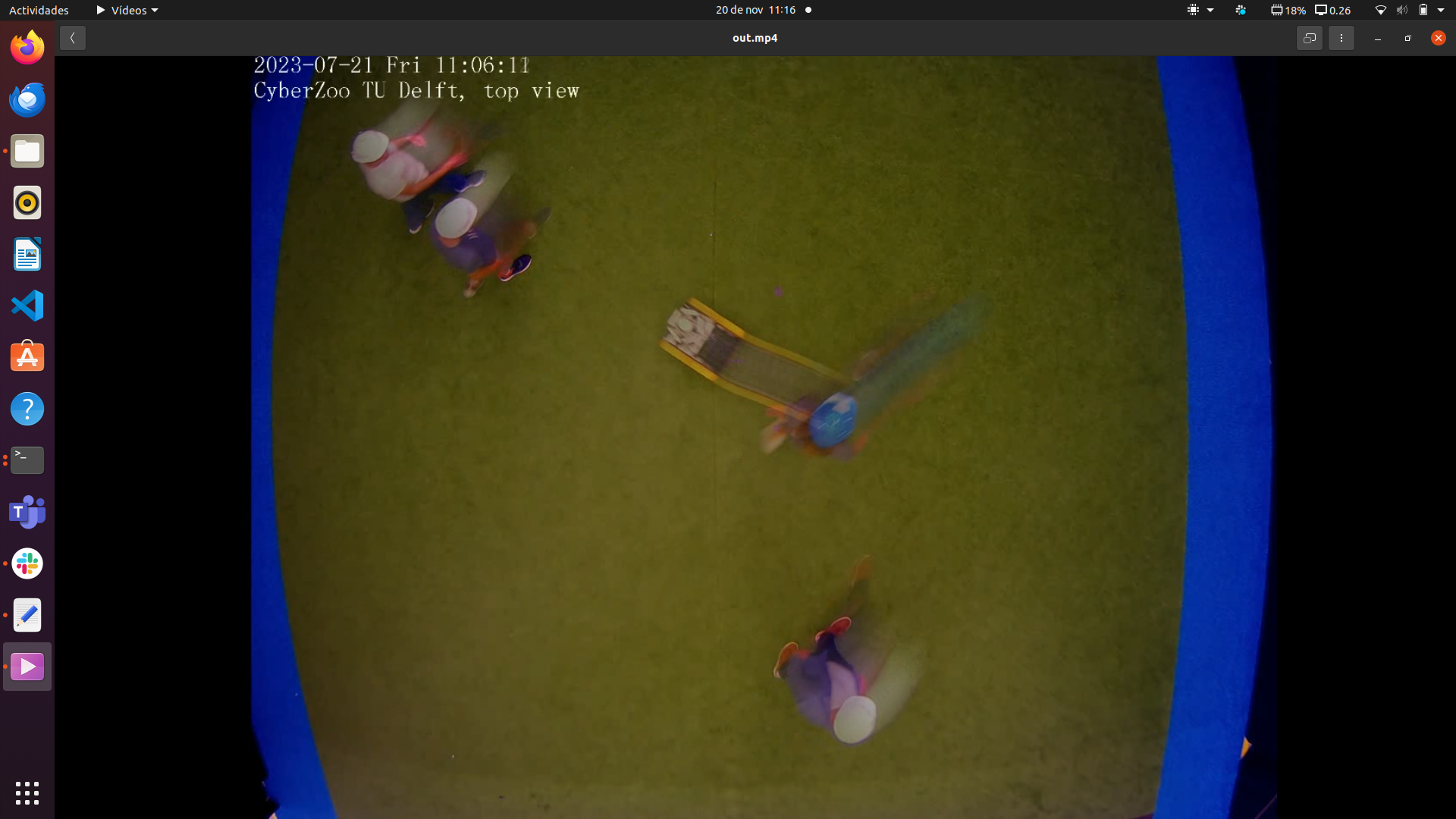}  & 
         \includegraphics[trim={22.5cm 2.9cm 10.5cm 6.75cm},clip,width=0.22\textwidth]{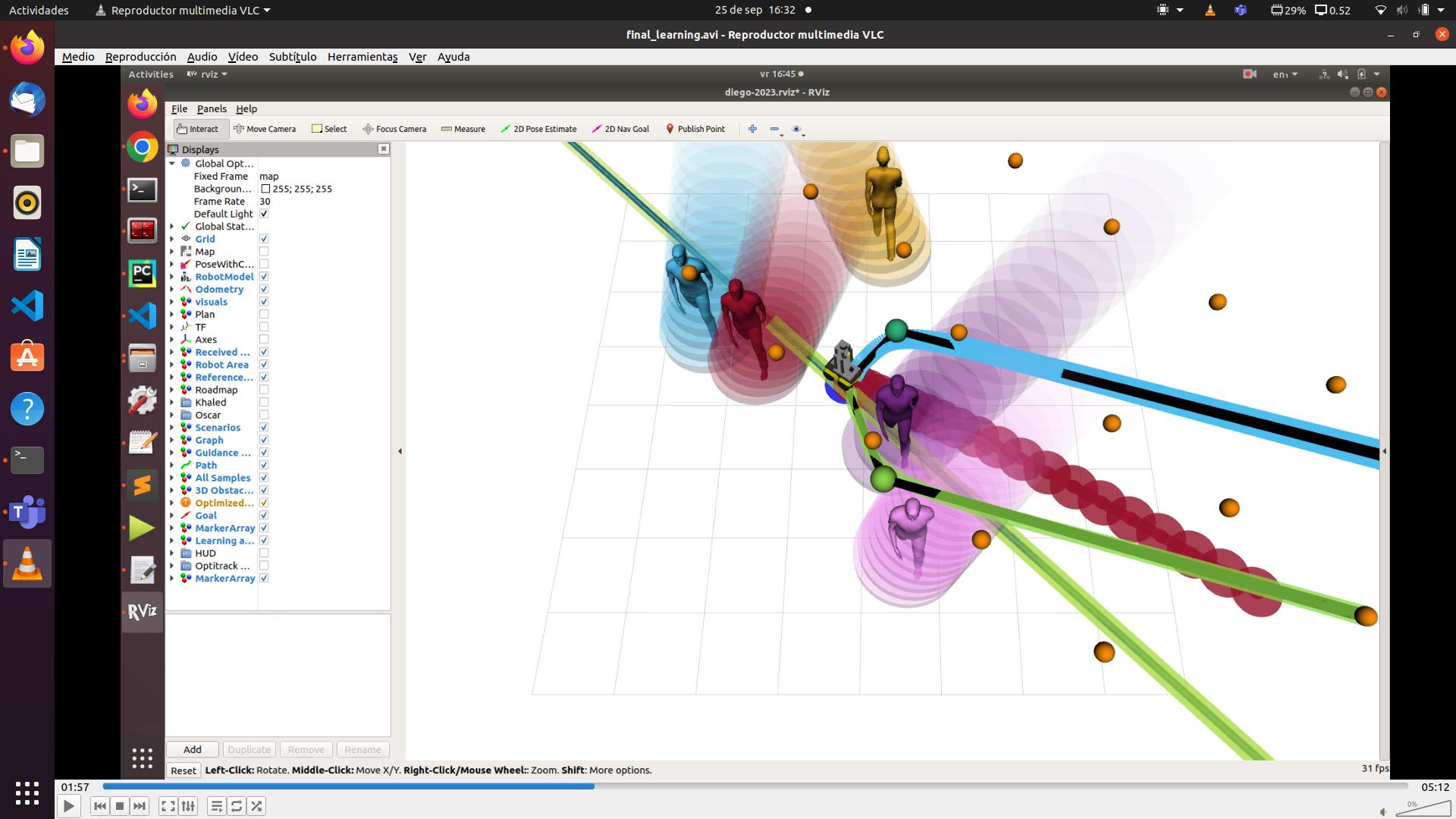}\\
         \footnotesize{(b)} & 
        \footnotesize{(e)} \\
        \includegraphics[trim={13.5cm 0.4cm 13.5cm 4.5cm},clip,width=0.22\textwidth]{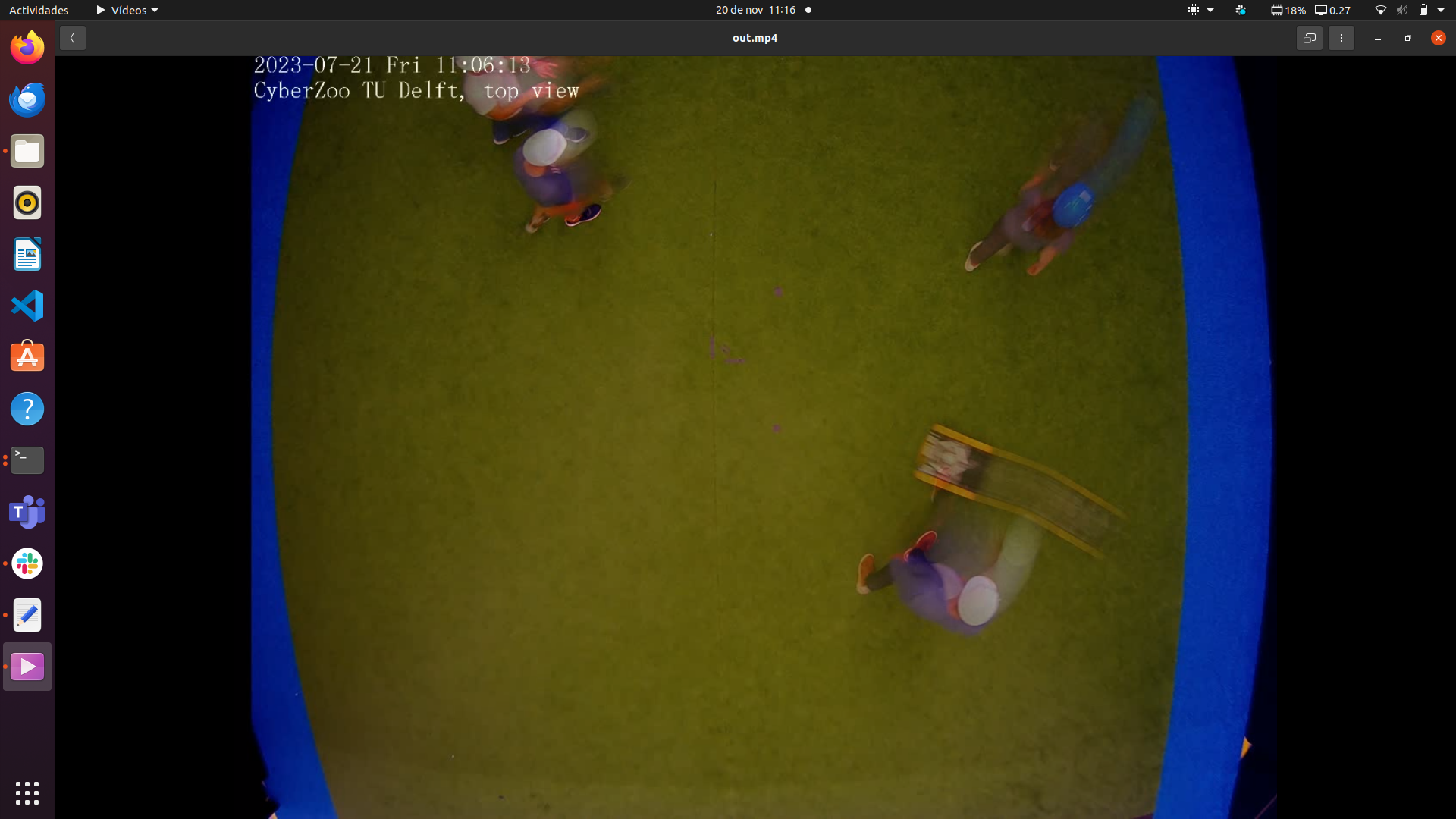}  & 
         \includegraphics[trim={22.5cm 2.9cm 10.5cm 6.75cm},clip,width=0.22\textwidth]{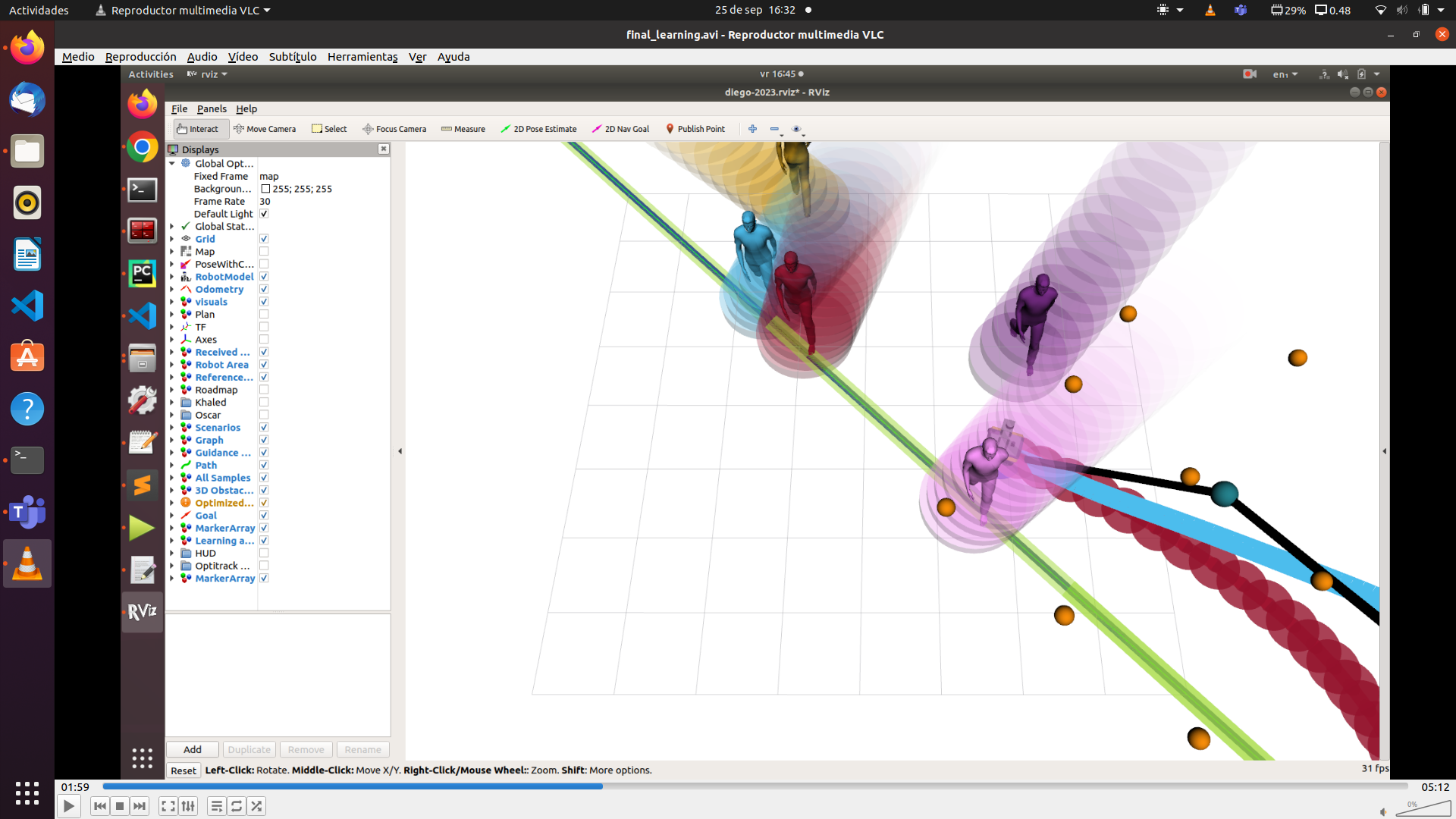}\\
         \footnotesize{(c)} & 
        \footnotesize{(f)} \\
    \end{tabular}
    \caption{Video frames and visualization of a sequence where the robot passes after a human with a high velocity and before a human with low velocity. The visualization is the same as in Figure~\ref{fig:change-homo}.}
    \label{fig:real-pass}
\end{figure}

Another example of interesting behavior is shown in Figure~\ref{fig:real-pass}. In that sequence, the robot faces a crossing with two humans at the same time, one of them with high velocity and the other one with low velocity. The robot chooses to pass after the pedestrian with high velocity, as its clear intention is passing as fast as possible, and to pass before the one with low velocity, who is not disturbed by the robot.

\section{Discussion} \label{sec:discussion}

The results of this work hold remarkable implications for the field of social navigation. One of the key implications is the ability to learn high-level discrete decisions from humans, unlike most of the learning-based social navigation approaches, which focus on imitating the low level behavior of humans (like DRL or imitation learning). They do not have collision constraints, and their performance is limited to \hh{the scenarios and behaviors seen during training, which may not exhibit all the intricacies of the real world, being extremely sensitive to out-of-distribution data}. Another problem that some learning-based approaches have is that they need a simulator to train the model, so their performance is limited to the realism of the simulations. With SHINE, however, in the worst case, the network will select a collision-free path that goes through a non-social homology class, still leading to a smooth and collision-free trajectory. Moreover, our work could be easily integrated with local optimization-based planners that include social costs in the optimization, achieving local social motion too. Therefore, our method is safer and easier to train compared to pure learning-based approaches, while we still incorporate learned social aspects to the navigation problem.

Our model has shown a remarkable adaptability to different environments and scenarios. In addition, the modularity design of the network allows the incorporation of more contextual information from the environment (as in Trajectron++ \citep{SalzmannIvanovicEtAl2020}). 

SHINE provides a combination of global and local planning in dynamic environments. Unlike only reactive and local planners, it may change between homology classes in real time. This is specially important in crowded environments, as human decisions may change in time and the robot behavior should adapt to those changes. 

\hh{The homology class selection module relies solely on the previous positions of obstacles and the robot and has been trained using real-world human trajectory data, implicitly accounting for possible interactions with humans and among humans themselves. Through this approach, SHINE implicitly couples prediction and planning, using the learned knowledge in the planning process. Although the local planner, LMPCC, does not integrate prediction and planning, it generates a local plan within that homology class, thereby accounting for human interactions at a global level.}

\hh{The prediction module does not try to find an optimal homology class, as it would depend on the user-specified navigation objective of the robot and it may even not exist. The goal of the module is finding the homology class that is the most likely to be chosen by a human. Due to the natural multi-modality in human trajectory selection, achieving a perfect prediction accuracy is impossible, having scenarios where more than one homology class are equally good depending on specific human preferences. Nevertheless, this is not a problem for SHINE, as selecting any equally good homology class results in equally good high-level motion planning.}

\section{Conclusion} \label{sec:conclusion}

This work presented a novel motion planner for dynamic environments that is capable of making human-like high-level decisions to navigate. To this end, we trained a model that predicts the difference between the homology class of a given trajectory and the homology class chosen by a human.
In particular, the planner first detects the possible homology classes of the trajectories in the environment. Then, it uses the trained prediction model to estimate the homology class that is the most likely to be chosen by a human, and navigates according to that decision. We showed that our planner learned social behaviors such as a tendency of passing behind other pedestrians or complex decisions to improve humans' comfort, and extensively evaluated it in both simulation and a lab experiment. We compared it with other social motion planners designed for social navigation and analyzed how it overcomes their limitations. In addition, it demonstrated a safe and smooth behavior in a crowded real-world environment. Furthermore, the prediction framework shows impressive results in predicting the homology class chosen by the humans, which could be used in other problems and open new research lines. 

This work could present certain limitations if it is applied in environments where the humans behavior considerably varies from the one in the training data; and it does not consider distinctions between robots and humans as dynamic obstacles or the underlying topology of the map. These limitations present opportunities for improvement by adding more training data of heterogeneous situations or integrating additional map or multi-robot features into the network architecture, due to its a modular design.

\hh{Even though SHINE accounts for human future interactions in a high level, the local planner used, LMPCC, does not plan considering them. Future work could involve studying how to integrate a local planner that couples low level interactions with local planning. In addition, extending the training data with datasets that involve other scenarios such as groups of people could expand and improve SHINE performance and adaptability with more kinds of social interactions.}

\begin{dci}
The authors declared no potential conflicts of interest with respect to the research, authorship, and/or publication of this article.
\end{dci}

\begin{funding}
The author(s) disclosed receipt of the following financial support
for the research, authorship, and/or publication of this article: This
work was supported by MICIU/AEI/10.13039/501100011033 and
ERDF/EU under grants (PID2022-139615OB-I00 and PRE2020-094415), from the Government of Aragón under grant DGA
T45-23R, from the European Union’s Horizon 2020 research and
innovation programme under grant agreement No. (101017008),
and from the European Union ERC, INTERACT, (101041863).
Views and opinions expressed are however those of the author(s)
only and do not necessarily reflect those of the European Union or
the European Research Council Executive Agency. Neither the
European Union nor the granting authority can be held responsible
for them.
\end{funding}

\bibliographystyle{SageH}
\bibliography{references}

\end{document}